\documentclass[lettersize,journal]{IEEEtran}

\usepackage{amsmath,amsfonts,bm}

\def\1{\bm{1}}

\def\vtheta{{\bm{\theta}}}
\def\va{{\bm{a}}}

\def\vo{{\bm{o}}}

\def\vu{{\bm{u}}}
\def\vv{{\bm{v}}}

\def\vx{{\bm{x}}}
\def\vy{{\bm{y}}}

\def\mA{{\bm{A}}}

\def\mI{{\bm{I}}}

\DeclareMathAlphabet{\mathsfit}{\encodingdefault}{\sfdefault}{m}{sl}
\SetMathAlphabet{\mathsfit}{bold}{\encodingdefault}{\sfdefault}{bx}{n}

\usepackage{amsmath,amsfonts}
\usepackage{array}
\usepackage[caption=false,font=normalsize,labelfont=sf,textfont=sf]{subfig}
\usepackage{textcomp}
\usepackage{stfloats}
\usepackage{url}
\usepackage{verbatim}
\usepackage{graphicx}
\usepackage{cite}
\usepackage{graphicx} % Required for inserting images
\usepackage{booktabs} % better visualization of table
\usepackage{multirow}
\usepackage[table]{xcolor}
\usepackage{colortbl}
\usepackage[ruled, linesnumbered, vlined]{algorithm2e}
\usepackage{longtable}
\usepackage{amsthm} % proof module
\usepackage{hyperref}
\usepackage{siunitx}  
\newcommand{\euclideanspace}{\mathbb{R}}
\newcommand{\manifold}{\mathcal{M} }
\newcommand{\tangentspace}[1]{$\mathcal{T}_{#1}\mathcal{M}$}
\newcommand{\expmap}[2]{\text{Exp}_{#1}(#2)}  % Exponential map of #2 at #1
\newcommand{\logmap}[2]{\text{Log}_{#1}(#2)}  % Logarithmic map of #2 at #1
\newtheorem{theorem}{Theorem}

\definecolor{darkgreen}{rgb}{0.0, 0.39, 0.0}
\definecolor{middlegreen}{rgb}{0.0, 0.5, 0.0}

\begin{document}

\title{Fast and Robust \\Visuomotor Riemannian Flow Matching Policy}
\author{Haoran Ding$^{1}$, Noémie Jaquier$^{2}$, Jan Peters$^{3}$, Leonel Rozo$^{1}$
        % <-this % stops a space
\thanks{$^{1}$Bosch Center for Artificial Intelligence. Renningen, Germany. \href{mailto:leonel.rozo@de.bosch.com}{\textrm{leonel.rozo@de.bosch.com}}}
\thanks{$^{2}$Division of Robotics, Perception, and Learning, KTH Royal Institute of Technology, Stockholm, Sweden. 
\href{mailto:jaquier@kth.se}{\textrm{jaquier@kth.se}}}
\thanks{$^{3}$ Computer Science Department of the Technische Universität Darmstadt, Darmstadt, Germany. 
\href{mailto:peters@ias.tu-darmstadt.de}{\textrm{peters@ias.tu-darmstadt.de}}}
\thanks{This work was partially supported by the Wallenberg AI, Autonomous Systems and Software Program (WASP) funded by the Knut and Alice Wallenberg Foundation.}
%\thanks{Manuscript received April 19, 2021; revised August 16, 2021.}
}

% The paper headers
\markboth{}%
{Ding \MakeLowercase{\textit{et al.}}: Fast and Robust Visuomotor Riemannian Flow Matching Policy}

\maketitle

\begin{abstract}
Diffusion-based visuomotor policies excel at learning complex robotic tasks by effectively combining visual data with high-dimensional, multi-modal action distributions. However, diffusion models often suffer from slow inference due to costly denoising processes or require complex sequential training arising from recent distilling approaches. 
This paper introduces Riemannian Flow Matching Policy (RFMP), a model that inherits the easy training and fast inference capabilities of flow matching (FM).
Moreover, RFMP inherently incorporates geometric constraints commonly found in realistic robotic applications, as the robot state resides on a Riemannian manifold.
To enhance the robustness of RFMP, we propose Stable RFMP (SRFMP), which leverages LaSalle's invariance principle to equip the dynamics of FM with stability to the support of a target Riemannian distribution. Rigorous evaluation on ten simulated and real-world tasks show that RFMP successfully learns and synthesizes complex sensorimotor policies on Euclidean and Riemannian spaces with efficient training and inference phases, outperforming Diffusion Policies and Consistency Policies.
\end{abstract}


\section{Introduction}
\IEEEPARstart{D}{eep} generative models are revolutionizing robot skill learning due to their ability to handle high-dimensional multimodal action distributions and interface them with perception networks, enabling robots to learn sophisticated sensorimotor policies~\cite{jurain2024deepsurvey}.
In particular, diffusion-based models such as diffusion policies (DP)~\cite{diffpol, consistencypolicy, goalconditionscorebasedpolicy,diffusionaugmentbc, 3ddiffpol, yang2024equibot} exhibit exceptional performance in imitation learning for a large variety of simulated and real-world robotic tasks, demonstrating a superior ability to learn multimodal action distributions compared to previous behavior cloning methods~\cite{florence2022implicit, bet, robomimic}. 
Nevertheless, these models are characterized by an expensive inference process as they often require to solve a stochastic differential equation, thus hindering their use in certain robotic settings~\cite{luo22:DiffModels}, e.g., for highly reactive motion policies. 

For instance, DP~\cite{diffpol}, typically based on a Denoising Diffusion Probabilistic Model (DDPM)~\cite{ho2020ddpm}, requires approximately $100$ denoising steps to generate an action. This translates to roughly $1$ second on a standard GPU. Even faster approaches such as Denoising Diffusion Implicit Models (DDIM)~\cite{song2020ddim} still need $10$ denoising steps, i.e., $0.1$ second, per action~\cite{diffpol}. Consistency policy~\cite{consistencypolicy} aims to accelerate the inference process by training a student model to mimic a DP teacher with larger denoising steps. Despite providing a more computationally-efficient inference, the CP training requires more computational resources and might be unstable due to the sequential training of the two models.
Importantly, training these models becomes more computationally demanding when manipulating data with geometric constraints, e.g., robot end-effector orientations, as the computation of the score function of the diffusion process is not as simple as in the Euclidean case~\cite{Huang22:RiemannianDM}. Furthermore, the inference process also incurs increasing computational complexity.

\begin{figure}[tb]
    \centering
    \includegraphics[trim={0.6cm 0.1cm 0.6cm 1.0cm}, clip, width=.32\columnwidth]{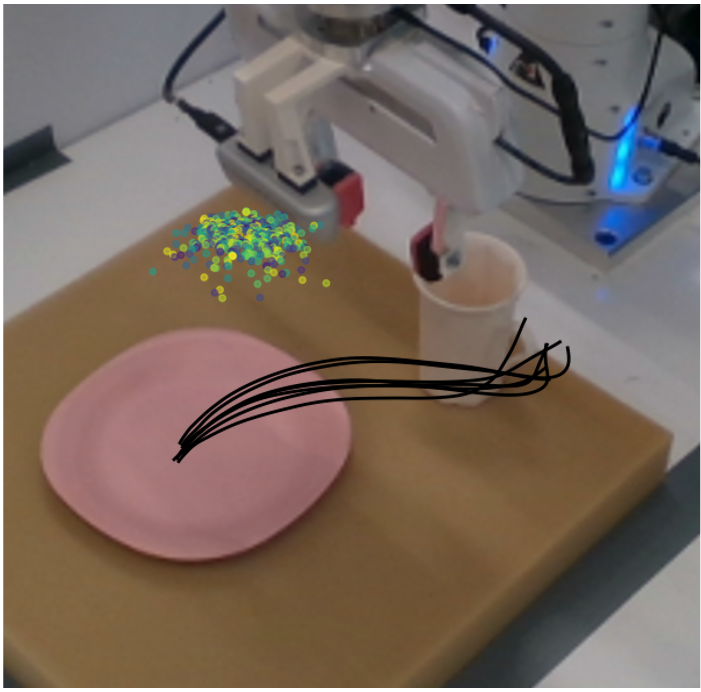}
    \includegraphics[trim={0.6cm 0.1cm 0.6cm 1.0cm}, clip, width=.32\columnwidth]{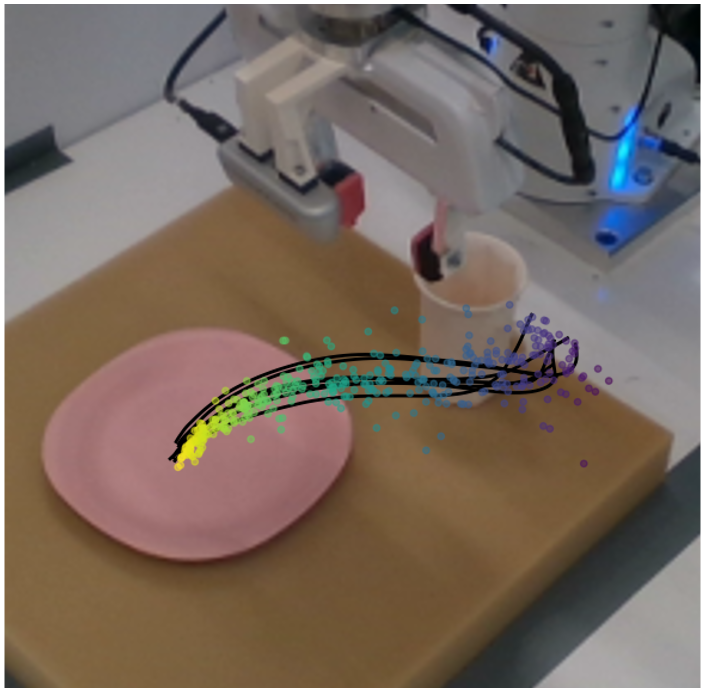}
    \includegraphics[trim={0.6cm 0.1cm 0.6cm 1.0cm}, clip, width=.32\columnwidth]{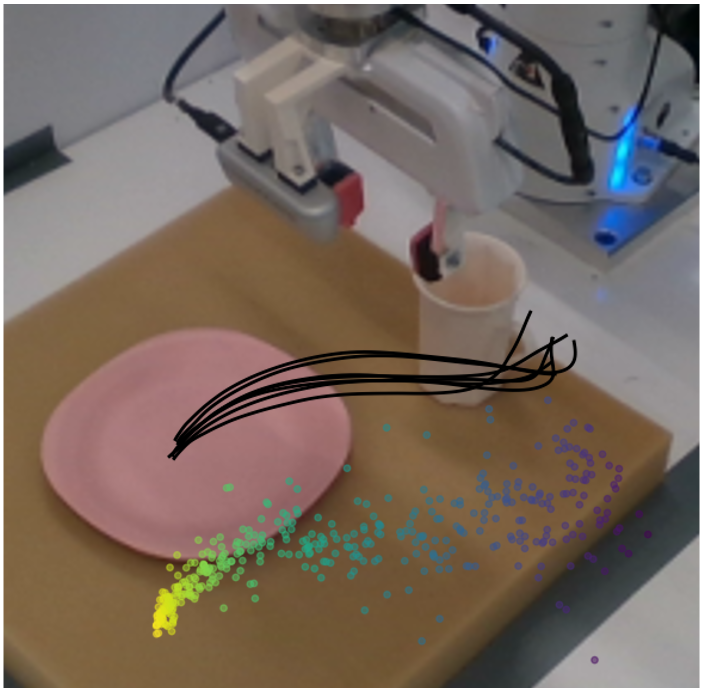}
    
    \vspace{0.1cm}
    
    \includegraphics[trim={0.6cm 0.1cm 0.6cm 1.0cm}, clip, width=.32\columnwidth]{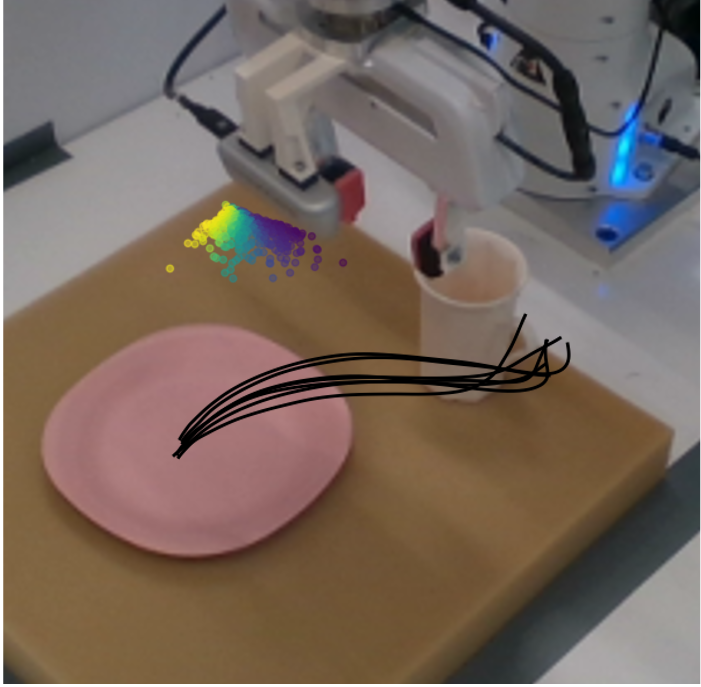}
    \includegraphics[trim={0.6cm 0.1cm 0.6cm 1.0cm}, clip, width=.32\columnwidth]{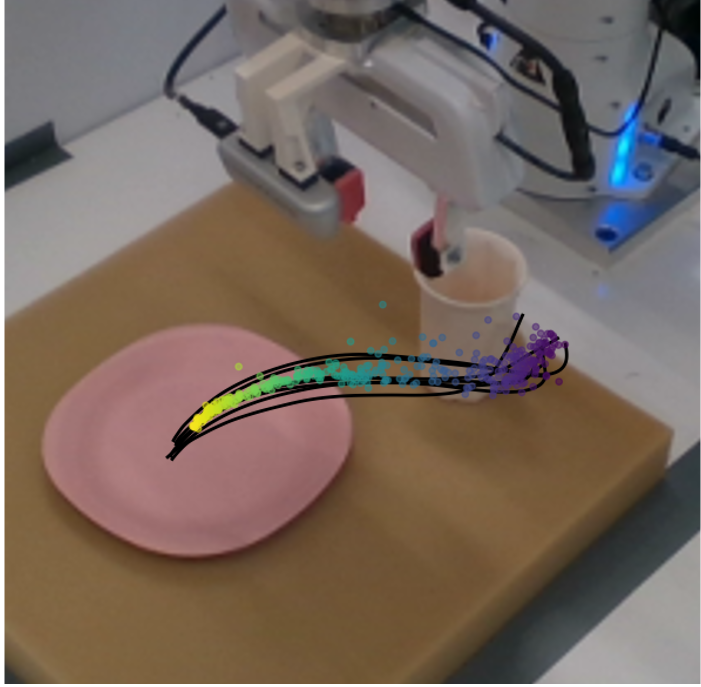}
    \includegraphics[trim={0.6cm 0.1cm 0.6cm 1.0cm}, clip, width=.32\columnwidth]{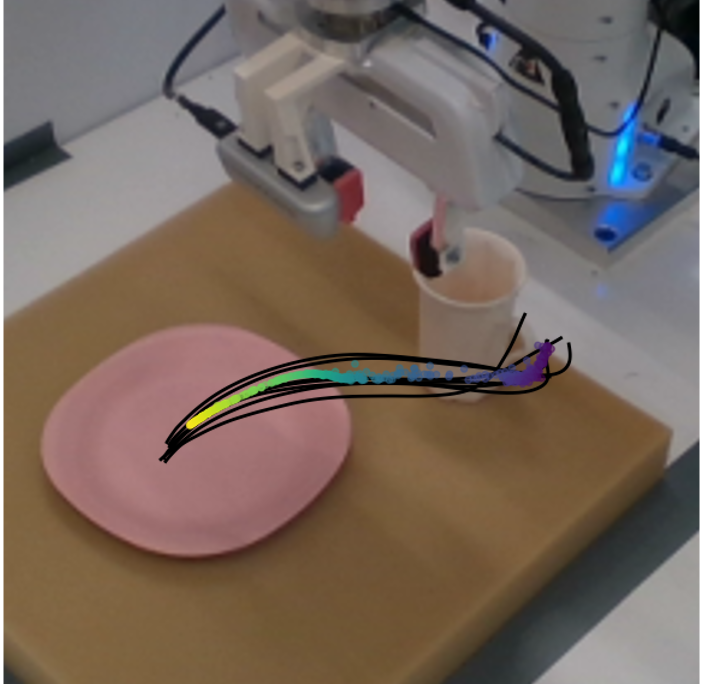}
    \caption{Flows of the RFMP (\textbf{top}) and SRFMP (\textbf{bottom}) at times ${t=\{0.0, 1.0, 1.5\}}$. The policies are learned from pick-and-place demonstration (black) and conditioned on visual observations. Note that the flow of SRFMP is stable to the target distribution at $t>1.0$, enhancing the policy robustness and inference time.}
    \label{fig:RFMP_overview}
\end{figure}

To overcome these limitations, we propose to learn visuomotor robot skills via a Riemannian flow matching policy (RFMP).
Compared to DP, RFMP builds on another kind of generative model: Flow Matching (FM)~\cite{lipmanCFM,riemannianfm}. Intuitively, FM gradually transforms a simple prior distribution into a complex target distribution via a vector field, which is represented by a simple function. The beauty of FM lies in its simplicity, as the resulting flow, defined by an ordinary differential equation (ODE), is much easier to train and much faster to evaluate compared to the stochastic differential equations of diffusion models.
However, as many visuomotor policies are represented in the robot's operational space, 
action representations must include both end-effector position and orientation. Thus, the policy must consider that orientations lie on either the $\mathcal{S}^{3}$ hypersphere or the $\operatorname{SO}(3)$ Lie group, depending on the chosen parametrization.
To properly handle such data, we leverage Riemannian flow matching (RFM)~\cite{riemannianfm}, an extension of flow matching that accounts for the geometric constraints of data lying on Riemannian manifolds.
In our previous work~\cite{rfmp}, we introduced the idea of leveraging flow matching in robot imitation learning and presented RFMP, which capitalizes on the easy training and fast inference of FM methods to learn and synthesize end-effector pose trajectories. However, our initial evaluation was limited to simple proof-of-concept experiments on the LASA dataset~\cite{Lemme2015:LasaDataset}.

\begin{figure}[tbp]
    \centering
    \includegraphics[width=0.95\linewidth]{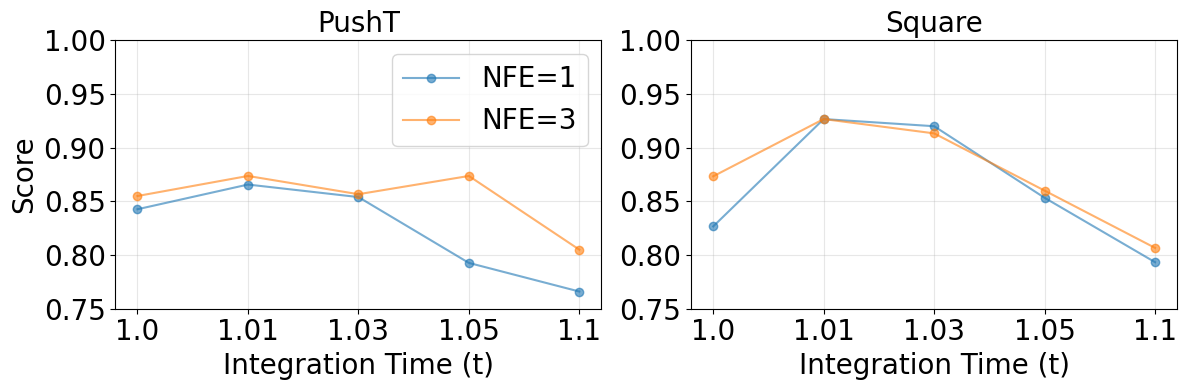}
    \caption{RFMP performance when extending inference time beyond $t=1$ on the Euclidean \textsc{PushT} and Robomimic \textsc{Square} tasks.}
    \label{figure:rfmp_beyond_t1}
\end{figure}

\textbf{In this paper}, we demonstrate the effectiveness of RFMP to learn complex real-world visuomotor policies and present a systematic evaluation of the performance of RFMP on both simulated and real-world manipulation tasks. Moreover, we propose Stable Riemannian Flow Matching Policy (SRFMP) to enhance the robustness of RFMP, as RFMP performance can be sensitive to the ODE integration horizon during inference (see Figures~\ref{fig:RFMP_overview} and~\ref{figure:rfmp_beyond_t1}). SRFMP builds on stable flow matching (SFM)~\cite{Sprague24:StableFM,stableflow}, which leverages LaSalle's invariance principle~\cite{LaSalle60:InvariancePrinciple} to equip the dynamics of FM with stability to the support of the target distribution. 
Unlike SFM, which is limited to Euclidean spaces, SRFMP generalizes this concept to Riemannian manifolds, guaranteeing the stability of the RFM dynamics to the support of a Riemannian target distribution. 
We systematically evaluate RFMP and SRFMP across $10$ tasks in both simulation and real-world settings, with policies conditioned on both state and visual observations. Our experiments demonstrate that RFMP and SRFMP inherit the advantages from FM models, achieving comparable performance to DP with fewer evaluation steps (i.e., faster inference) and significantly shorter training times. Moreover, 
under the same training epochs, our methods outperform both DP and CP. 
Notably, SRFMP requires fewer ODE steps than RFMP to achieve an equivalent performance, resulting in even faster inference times.

In summary, beyond demonstrating the effectiveness of our early work on simulated and real robotic tasks, \textbf{the main contributions of this article} are threefold: \emph{(1)} We introduce Stable Riemannian Flow Matching (SRFM) as an extension of SFM~\cite{stableflow} to incorporate stability into RFM; \emph{(2)} We propose stable Riemannian flow matching policy (SRFMP), which combines the easy training and fast inference of RFMP with stability guarantees to a Riemannian target action distribution; \emph{(3)} We systematically evaluate both RFMP and SRFMP across $10$ tasks from simulated benchmarks and real settings. Supplementary material is available on the paper website \url{https://sites.google.com/view/rfmp}.

\section{Related work}
\label{sec:related_work}
As the literature on robot policy learning is vast, we here focus on approaches that design robot policies based on flow-based generative models.

\textbf{Normalizing Flows} are arguably the first broadly-used generative models in robot policy learning~\cite{normalizingflow}. They were commonly employed as diffeomorphisms for learning stable dynamical systems in Euclidean spaces~\cite{euclideanflowstabledynsys, learningstablenfcontrol, Imitationflow}, with extensions to Lie groups~\cite{urain2022learning} and Riemannian manifolds~\cite{zhang2022learning}, similarly addressed in this paper. The main drawback of normalizing flows is their slow training, as the associated ODE needs to be integrated to calculate the model log-likelihood. Moreover, none of the foregoing works learned sensorimotor policies based on visual observations via imitation learning.

\textbf{Diffusion Models}~\cite{Yang23:DiffModelsSurvey} recently became state-of-art in imitation learning due to their ability to learn multi-modal and high-dimensional action distributions. They have been primarily employed to learn motion planners~\cite{janner2022planningdiffusion}, and complex control policies~\cite{diffpol,Wang23:DiffPolORL,goalconditionscorebasedpolicy}.
Recent extensions use 3D visual representations from sparse point clouds~\cite{3ddiffpol}, and employ equivariant networks for learning policies that, by design, are invariant to changes in scale, rotation, and translation~\cite{yang2024equibot}. However, a major drawback of diffusion models is their slow inference process. In~\cite{diffpol}, DP requires $10$ to $100$ denoising steps, i.e., $0.1$ to $1$ second on a standard GPU, to generate each action. Consistency models (CM)~\cite{consistencymodel} arise as a potential solution to overcome this drawback~\cite{consistencypolicy, manicm,onestepDP}. CM distills a student model from a pretrained diffusion model (i.e., a teacher), enabling faster inference by establishing direct connections between points along the probability path. Nevertheless, this sequential training process increases the overall complexity and time of the whole training phase.\looseness=-1

\textbf{Flow Matching}~\cite{lipmanCFM} essentially trains a normalizing flow by regressing a conditional vector field instead of maximizing the likelihood of the model, thus avoiding to simulate the ODE of the flow. This leads to a significantly simplified training procedure compared to classical normalizing flows. 
Moreover, FM builds on simpler probability transfer paths than diffusion models, thus facilitating faster inference. Tong \emph{et al.}~\cite{torchcfm} showed that several types of FM models can be obtained according to the choice of conditional vector field and source distributions, some of them leading to straighter probability paths, which ultimately result in faster inference.
Rectified flows~\cite{rectifiedflow} is a similar simulation-free approach that designs the vector field by regressing against straight-line paths, thus speeding up inference. Note that rectified flows are a special case of FM, in which a Dirac distribution is associated to the probability path~\cite{torchcfm}. 
Due to their easy training and fast inference, FM models quickly became one of the de-facto generative models in machine learning and have been employed in a plethora of different applications~\cite{se3fmprotein,sampleobservation,Hu23:MotionFlowMatching,Zhang24:TrajectoryFM,Lin24:PPFlow,Liu24:FMspeech}.\looseness=-1  

In our previous work~\cite{rfmp}, we proposed to leverage RFM to learn sensorimotor robot policies represented by end-effector pose trajectories on Riemannian manifolds. Building on a similar idea, subsequent works have used FM along with an equivariant transformer to learn $\operatorname{SE}(3)$-equivariant policies~\cite{actionflow}, for multi-support manipulation tasks with a humanoid robot~\cite{humanoidimitflow}, and for robot imitation learning with point cloud observations~\cite{3dcfmp}. \textbf{In this paper}, we build upon our previous work, Riemannian Flow Matching Policy (RFMP)~\cite{rfmp}, to enable the learning of complex visuomotor policies on Riemannian manifolds. Unlike the aforementioned approaches, our work focuses on providing a fast and robust RFMP inference process. We achieve this by constructing the FM vector field using LaSalle's invariance principle, which not only enhances inference robustness with stability guarantees but also preserves the easy training and fast inference capabilities of RFMP.

\section{Background}
\label{section: background}
In this section, we provide a short background on Riemannian manifolds, and an overview of the flow matching framework with its extension to Riemannian manifolds. 

% - Riemannian manifolds
\subsection{Riemannian Manifolds}
\label{subsec:Riemannian Manifolds}
A smooth manifold $\manifold$ can be intuitively understood as a $d$-dimensional surface that locally, but not globally, resembles the Euclidean space $\euclideanspace^d$~\cite{DoCarmo92:RiemannianGeometry,lee2018RMintroduction}. The geometric structure of the manifold is described via the so-called charts, which are diffeomorphic maps between parts of $\manifold$ and $\euclideanspace$. The collection of these charts is called an atlas. 
The smooth structure of $\manifold$ allows us to compute derivatives of curves on the manifolds, which are tangent vectors to $\manifold$ at a given point $\vx$. 
For each point $\vx \in \manifold$, the set of tangent vectors $\vu$ of all curves that pass through $\vx$ forms the tangent space \tangentspace{\vx}. The tangent space spans a $d$-dimensional affine subspace of $\euclideanspace^d$, where $d$ is the manifold dimension. The collections of all tangent spaces of $\manifold$ forms the tangent bundle ${\mathcal{TM} = \bigcup_{\vx \in \manifold} \{(\vx, \vu)|\vu \in \mathcal{T}_\vx\manifold\}}$, which can be thought as the union of all tangent spaces paired with their corresponding points on $\manifold$. 

Riemannian manifolds are smooth manifolds equipped with a smoothly-varying metric $g$, which is a family of inner products $g_\vx: \mathcal{T}_\vx\manifold \times \mathcal{T}_\vx\manifold \to \mathbb{R}$. 
The norm associated with the metric is denoted as $\| \vv\|_{g_{\vx}}$ with $\vv\in \mathcal{T}_\vx\manifold$, and the distance between two vectors $\vu, \vv \in \mathcal{T}_\vx\manifold$ is defined as the norm $\Arrowvert\vu-\vv\Arrowvert_{g_{\vx}}$.
With this metric, we can then define the length of curves on $\manifold$. The shortest curve on $\manifold$ connecting any two points $\vx, \vy \in\manifold$ is called a geodesic. Intuitively, geodesics can be seen as the generalization of straight lines to Riemannian manifolds. To operate with Riemannian manifolds, a common way is to exploit its Euclidean tangent spaces $\mathcal{T}_\vx\manifold$ and back-and-forth maps between $\manifold$ and $\mathcal{T}_\vx\manifold$, i.e., the exponential and logarithmic maps. Specifically, the exponential map $\expmap{\vx}{\vu}: \mathcal{T}_\vx\manifold \to \manifold$ maps a point $\vu$ on the tangent space of $\vx$ to a point $\vy\in\manifold$, so that the geodesic distance between $\vy= \expmap{\vx}{\vu}$ and $\vx$ satisfies $d_{g}(\vx, \vy) = \| \vu \|_{g_{\vx}}$. The inverse operation is the logarithmic map $\logmap{\vx}{\vy}: \manifold \to \mathcal{T}_\vx\manifold$, which projects a point $\vy\in\manifold$ to the tangent space $\mathcal{T}_\vx\manifold$ of $\vx$. Finally, when optimizing functions of manifold-valued parameters, we need to compute the Riemannian gradient. Specifically, the Riemannian gradient of a scalar function $f: \manifold \to \mathbb{R}$ at $\vx\in\manifold$ is a vector in the tangent space $\mathcal{T}_\vx\manifold$~\cite{Absil07:RiemannOpt,Boumal22:RiemannOpt}. It is obtained via the identification $\mathcal{L}_{\vu} f(\vx)= \langle \nabla_{\vx} f(\vx), \vu \rangle_{\vx}$, where $\mathcal{L}_{\vu} f(\vx)$ denotes the directional derivative of $f$ in the direction $\vu\in\mathcal{T}_\vx\manifold$, and $\langle \cdot, \cdot \rangle_{\vx}$ is the Riemannian inner product on $\mathcal{T}_\vx\manifold$.

% - Flow matching
\subsection{Flow Matching}
\label{subsec:FM}
Continuous normalizing flows (CNF)~\cite{neuralode} form a class of deep generative models that transform a simple probability distribution into a more complex one. The continuous transformation of the samples is parametrized by an ODE, which describes the flow of the samples over time. Training CNF is achieved via maximum likelihood estimation, and thus involves solving (a.k.a. simulating) inverse ODEs, which is computationally expensive. Instead, flow matching (FM)~\cite{lipmanCFM} is a simulation-free generative model that efficiently trains CNF by directly mimicking a target vector field.

\subsubsection{Euclidean Flow Matching}
FM~\cite{lipmanCFM} reshapes a simple prior distribution $p$ into a (more complicated) target distribution $q$ via a probability density path $p_t$ that satisfies $p_0=p$ and $p_1=q$. The path $p_t$ is generated by push-forwarding $p_0$ along a flow $\psi_t$ as, 
 \begin{equation}
     \label{equation:psuhfroward}
     p_t=[\psi_t]_* p_0,
 \end{equation}
where the push-forward operator $*$ is defined as,
 \begin{equation}
     [\psi_t]_*p_0(\vx) = p_0(\psi_t^{-1}(\vx))\det\left(\frac{\partial\psi_t^{-1}(\vx)}{\partial \vx}\right).
 \end{equation}
The flow $\psi_t$ is defined via a vector field ${u_t: [0, 1] \times \euclideanspace^d \to \euclideanspace^d}$ by solving the ODE, 
\begin{equation}
    \label{equation:flow}
    \frac{d\psi_t(\vx)}{dt} = u_t(\psi_t(\vx)), \quad \text{with} \quad \psi_0(\vx) = \vx.
\end{equation}

Assuming that both the vector field $u_t(\vx)$ and probability density path $p_t$ are known, one can regress a parametrized vector field $v_t(\vx; \vtheta): [0, 1] \times \euclideanspace^d \to \euclideanspace^d$ to some target vector field $u_t$, which leads to the FM loss function,
\begin{equation}
    \ell_{\text{FM}}(\vtheta) = \mathbb{E}_{t,p_t(\vx)}\Arrowvert v_t(\vx;\vtheta) - u_t(\vx)\Arrowvert _2^{2},
    \label{equation:FMloss}
\end{equation}
where $\vtheta$ are the learnable parameters, $t \sim \mathcal{U}[0, 1]$, and $\vx\sim p_t(\vx)$. However, the loss~\eqref{equation:FMloss} is intractable since we actually do not have prior knowledge about $u_t$ and $p_t$. Instead, Lipman \emph{et al.}~\cite{lipmanCFM} proposed to learn a conditional vector field $u_t(\vx|\vx_1)$ with $\vx_1$ as a conditioning variable. This conditional vector field generates the conditional probability density path $p_t(\vx|\vx_1)$, which is related to the marginal probability path via $p_t(\vx)=\int p_t(\vx|\vx_1)q(\vx_1)d\vx_1$, with $q$ being the unknown data distribution. After reparametrization, this leads to the tractable conditional flow matching (CFM) loss function,
\begin{equation}
    \ell_{\text{CFM}}(\vtheta)=\mathbb{E}_{t, q(\vx_1), p(\vx_0)} \Arrowvert v_t(\vx_t;\vtheta) - u_t(\vx_t|\vx_1) \Arrowvert _2^{2},
    \label{equation:CFMloss}
\end{equation}
where $\vx_t=\psi_t(\vx_0|\vx_1)$ denotes the conditional flow.
Note that optimizing the CFM loss~\eqref{equation:CFMloss} is equivalent to optimizing the FM loss~\eqref{equation:FMloss} as they have identical gradients~\cite{lipmanCFM}. 
The problem therefore boils down to design a conditional vector field $u_t(\vx_t|\vx_1)$ that generates a probability path $p_t$ satisfying the boundary conditions $p_0=p$, $q=p_1$. Intuitively, $u_t(\vx_t|\vx_1)$ should move a randomly-sampled point at $t=0$ to a datapoint at $t\!=\!1$.
Lipman \emph{et al.}~\cite{lipmanCFM} proposed the Gaussian CFM, which defines a probability path from a zero-mean normal distribution to a Gaussian distribution centered at $\bm{x}_1$ via the conditional vector field,
\begin{equation}
    u_t(\vx_t|\vx_1) = \vx_1 - (1-\sigma)\vx_0 , 
    \label{Eq:LipmanCFM}
\end{equation}
which leads to the flow $\vx_t = \psi_t(\vx_0|\vx_1) = (1-(1-\sigma)t)\vx_0 + t\vx_1$. Note that a more general version of CFM is proposed by Tong \emph{et al.}~\cite{torchcfm}.

Finally, the inference process of CFM is straightforward and consists of the following steps: \emph{(1)} Get a sample from $p_0$; and \emph{(2)} Query the learned vector field $v_t(\vx;\vtheta)$ to solve the ODE~\eqref{equation:flow} with off-the-shelf solvers, e.g., based on the Euler method~\cite{eulermethod}. 

\subsubsection{Riemannian Flow Matching}
In many robotics settings, data lies on Riemannian manifolds~\cite{robotdatageometry, introrobot}. For example, various tasks involve the rotation of the robot's end-effector. Therefore, the corresponding part of the state representation lies either on the Riemannian hypersphere $\mathcal{S}^{3}$ or the $\operatorname{SO}(3)$ group, depending on the choice of parametrization. To guarantee that the FM generative process satisfies manifold constraints, Chen and Lipman~\cite{riemannianfm} extended CFM to Riemannian manifolds. The Riemannian conditional flow matching (RCFM) considers that the flow $\psi_t$ evolves on a Riemannian manifold $\manifold$. Thus, for each point $\vx\in\manifold$, the vector field associated to the flow $\psi_t$ at this point lies on the tangent space of $\vx$, i.e., $u_t(\vx) \in \mathcal{T}_\vx\manifold$. The RCFM loss function resembles that of the CFM model but it is computed with respect to the Riemannian metric $g_{\vx}$ as follows,
\begin{equation}
    \ell_{\text{RCFM}}=\mathbb{E}_{t, q(\vx_1), p(\vx_0)} \Arrowvert v_t(\vx_t;\vtheta) - u_t(\vx_t|\vx_1)\Arrowvert _{g_{\vx_t}}^{2}.
    \label{equation:RCFMloss}
\end{equation}
As in the Euclidean case, we need to design the flow $\psi_t$, its corresponding conditional vector field $u_t(\vx_t|\vx_1)$, and choose the base distribution. 
Following~\cite{riemannianfm,se3fmprotein}, the most straightforward strategy is to exploit geodesic paths to design the flow $\psi_t$.
For simple Riemannian manifolds such as the hypersphere, the hyperbolic manifold, and some matrix Lie groups, geodesics can be computed via closed-form solutions. We can then leverage the geodesic flow given by,
\begin{equation}
    \label{equation:rfmflow}
     \vx_t = \expmap{\vx_1}{t \, \logmap{\vx_1}{\vx_0}}, \quad t \in [0,1].
\end{equation}
The conditional vector field can then be calculated as the time derivative of $\vx_t$, i.e., $u_t(\vx_t|\vx_1)=\dot{\vx}_t$. Notice that $u_t$ boils down to the conditional vector field~\eqref{Eq:LipmanCFM} with $\sigma=0$ when $\manifold=\mathbb{R}^d$. Chen and Lipman~\cite{riemannianfm} also provide a general formulation of $u_t(\vx_t|\vx_1)$ for cases where closed-form geodesics are not available.
The prior distribution $p_0$ can be chosen as a uniform distribution on the manifold~\cite{se3fmprotein,riemannianfm}, or as a Riemannian~\cite{Pennec06:RiemannianStats} or wrapped Gaussian~\cite{riemannianfm,wrappedgaussian} distribution on $\manifold$. During inference, we solve the corresponding RCFM's ODE on the Riemannian manifold $\manifold$ via projection-based methods. Specifically, at each step, the integration is performed in the tangent space $\mathcal{T}_\vx\manifold$ and the resulting vector is projected onto the Riemannian manifold $\manifold$ with the exponential map.

\subsection{Flow Matching vs. Diffusion and Consistency Models}
Diffusion Policy (DP)~\cite{diffpol} is primarily trained based on DDPM~\cite{ho2020ddpm}, which performs iterative denoising from an initial noise sample $\vx_K$, where $K$ denotes the total number of denoising steps. The denoising process follows,
\begin{equation}
    \vx_{k-1} = \alpha_t\big(\vx_{k} - \gamma\epsilon_{\bm{\theta}}(\vx_k, k) + \mathcal{N}(0, \sigma^2 \bm{I})\big),
\end{equation}
where $\alpha$, $\gamma$ and $\sigma$ constitute the noise schedule that governs the denoising process, and $\epsilon_{\theta}(\vx_k, k)$ is the noise prediction network that infers the noise at the $k$-th denoising step. The final sample $\vx_0$ is the noise-free target output. The equivalent inference process in FM is governed by,
\begin{equation}
    \label{equation:learned_flow}
    \frac{d\psi_t(\vx)}{dt} = v_t(\psi_t(\vx); \vtheta), \quad \text{with} \quad \psi_0(\vx) = \vx ,
\end{equation}
where $v_t(\psi_t(\vx); \vtheta)$ is the learned vector field that mimics the target forward process~\eqref{Eq:LipmanCFM}. 
Two key differences can be identified: (1) FM requires to solve a simple ODE, and (2) The FM vector field induces straighter paths.
Importantly, the DP denoising framework limits its inference efficiency, particularly in scenarios requiring faster predictions.

Consistency Policy (CP)~\cite{consistencypolicy} aims to speed up the inference process of DP by leveraging a consistency model that distills the DP as a teacher model. While CP adopts a similar denoising mechanism as DP, it enhances the process by incorporating both the current denoising step $k$ and the target denoising step $t$ as inputs to the denoising network, formalized as $\epsilon_{\theta}(\vx_k, k, t)$. However, CP involves a two-stage training process. First, a DP teacher policy is trained. Next, a student policy is trained to mimic the denoising process of the teacher policy. This approach enables CP to achieve faster and more efficient inference while retaining the performance of its teacher model, at the cost of a more complex training process. In contrast, FM training is simulation-free, and features a single-phase training with a simple loss function. 

\section{Fast and Robust RFMP}
\label{sec:FSRFMPs}
Our goal is to leverage the RCFM framework to learn a parameterized policy $\pi_\vtheta(\va|\vo)$ that adheres to the target (expert) policy $\pi_e(\va|\vo)$, which generates a set of $N$ demonstrations $D_n=\{\vo^s, \va^s \}_{s=1}^T$, where $\vo$ denotes an observation, $\va$ represents the corresponding action, and $T$ denotes the length of $n$-th trajectory. 
In this section, we first introduce Riemaniann flow matching polices (RFMP) that leverage RCFM to achieve easy training and fast inference. Second, we propose Stable RFMP (SRFMP), an extension of RFMP that enhances its robustness and inference speed through stability to the target distribution.

\subsection{Riemannian Flow Matching Policy}
\label{subsec:RFMP}
% Description of RFMPs, IROS2024
RFMP adapts RCFM to visuomotor robot policies by learning an observation-conditioned vector field $u_t(\va|\vo)$. Similar to DP~\cite{diffpol}, RFMP employs a receding horizon control strategy~\cite{recdinghorizoncontrol}, by predicting a sequence of actions over a prediction horizon $T_p$. This strategy aims at providing temporal consistency and smoothness on the predicted actions. This means that the predicted action horizon vector is constructed as $\va=[\va^s, \va^{s+1},\ldots, \va^{s+T_p}]$, where $\va^{i}$ is the action at time step $i$, and $T_p$ is the action prediction horizon. This implies that all samples $\va_1$, drawn from the target distribution $p_1$, have the form of the action horizon vector $\va$.
Moreover, we define the base distribution $p_0$ such that samples $\bm{a}_{p_0}\sim p_0$ are of the form $\bm{a}_0=[\bm{a}_{b}, \ldots, \bm{a}_{b}]$ with $\bm{a}_{b}$ sampled from an auxiliary distribution $b$. This structure contributes to the smoothness of the predicted action vector $\bm{a}$, as the flow of all its action components start from the same initial action $\bm{a}_{b}$. 

In contrast to the action horizon vector, the observation vector $\vo$ is not defined on a receding horizon but is constructed by randomly sampling only few observation vectors. Specifically, RFMP follows the sampling strategy proposed in~\cite{sampleobservation} which uses: \emph{(1)} A reference observation $\vo^{s-1}$ at time step $s-1$; \emph{(2)} A context observation $\vo^c$ randomly sampled from an observation window with horizon $T_o$, i.e., $c$ is uniformly sampled from $[s - T_o, s-2]$; and \emph{(3)} The time gap $s-c$ between the context observation and reference observation. Therefore, the observation vector is defined as $\vo = [\vo^{s-1}, \vo^c, s - c]$. Notice that, when $T_o = 2$, we disregard the time gap and the observation is $\vo = [\vo^{s-1}, \vo^c]$. 
The aforementioned strategy leads to the following RFMP loss function,
\begin{equation}
    \label{equation: rfmploss}
    \ell_{\text{RFMP}} = \mathbb{E}_{t, q(\va_1), p(\va_0)}\Arrowvert v_t(\va_t|\vo; \vtheta)-u_t(\va_t|\va_1)\Arrowvert_{g_{\va_t}}^{2}.
\end{equation}
Algorithm \ref{algorithm:rfmp training} summarizes the training process of RFMP. Note that our RFMP inherits most of the training framework of RCFM, the main difference being that the vector field learned in RFMP is conditioned on the observation vector $\vo$.
\begin{algorithm}[tb]
    \label{algorithm:rfmp training}
    \caption{RFMP Training \& Inference}
    \textbf{Training}\\
    \KwIn{Initial parameters $\vtheta$, prior and target distribution $p_0$, $p_1$.}
    \KwOut{Learned vector field parameters $\vtheta$.}
    \While{termination condition unsatisfied}{
        Sample flow time step $t$ from the uniform distribution $\mathcal{U}[0 ,1]$. \\
        Sample noise $\va_0 \sim p_0$. \\
        Jointly sample action sequence $\va_1\sim p_1$ and corresponding observation vector $\vo$. \\
        Compute conditional vector field $u_t(\va_t|\va_1)$ via the RCFM geodesic flow~\eqref{equation:rfmflow}.\\
        Evaluate $\ell_{\text{RFMP}}$ as in~\eqref{equation: rfmploss}.\\
        Update parameters $\vtheta$.
        }
    \textbf{Inference Step}\\
    \KwIn{Predefined number of function evaluation $N$, learned vector field $v_{\vtheta}$, observation vector $\vo$, prior distribution $p_0$.}
    Sample $\va_0 \sim p_0$, and set $t=0$.\\
    \While{$t \leq 1$}{
        Integrate the learned Riemannian vector field $\bm{\xi}_{t+\Delta t} = \text{Exp}_{\bm{\xi}_{t}} (v_{\vtheta}(\bm{\xi}_{t}, \vo) \Delta t)$. \\
        Update time $t = t + \Delta t$. \\
    }
\end{algorithm}
After training RFMP, the inference process, which essentially queries the policy $\pi_\vtheta(\va|\vo)$, boils down to the following four steps: \emph{(1)} Draw a sample $\va_0$ from the prior distribution $p_0$; \emph{(2)} Construct the observation vector $\vo$; \emph{(3)} Employ an off-the-shelf ODE solver to integrate the learned vector field $v_t(\va|\vo;\vtheta)$ from $\va_0$ along the time interval $t=[0, 1]$, and get the generated action sequence $\va = [\va^s,...,\va^{s+T_p}]$; and \emph{(4)} Execute the first $T_a$ actions of the sequence $\va$ with $T_a \le T_p$. This last step allows the robot to quickly react to environment changes, while still providing smooth predicted actions.

\begin{figure}[t]
\centering
\includegraphics[width=0.32\linewidth]{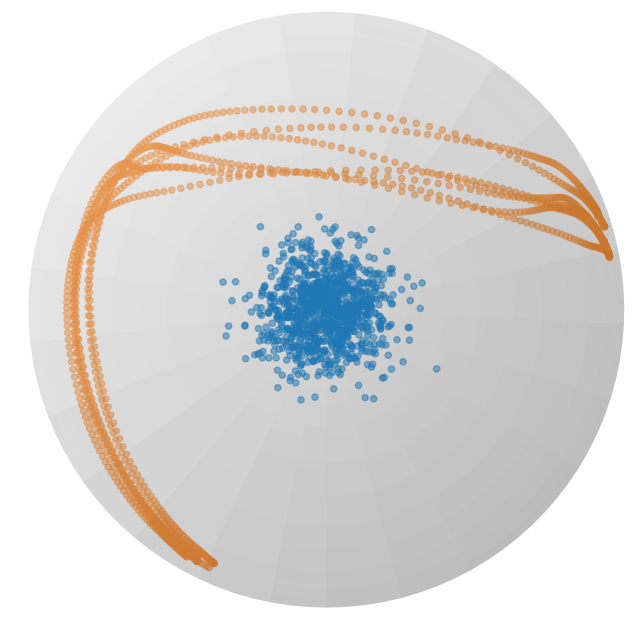} 
\includegraphics[width=0.32\linewidth]{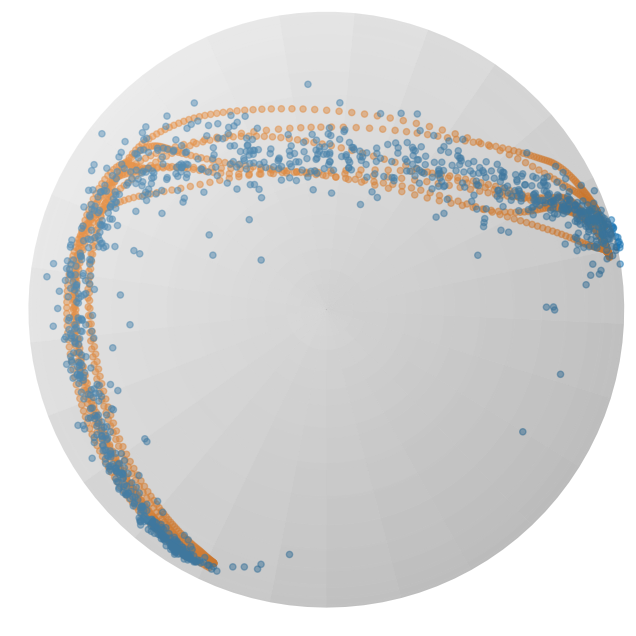}
\includegraphics[width=0.32\linewidth]{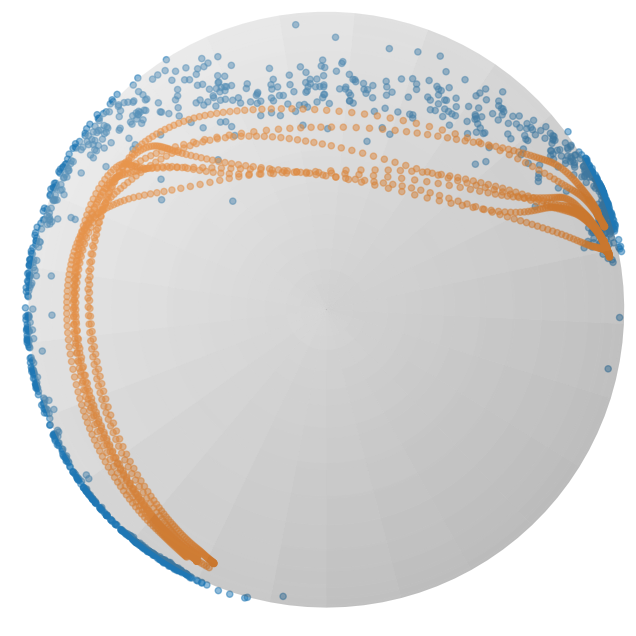} \\
\includegraphics[width=0.32\linewidth]{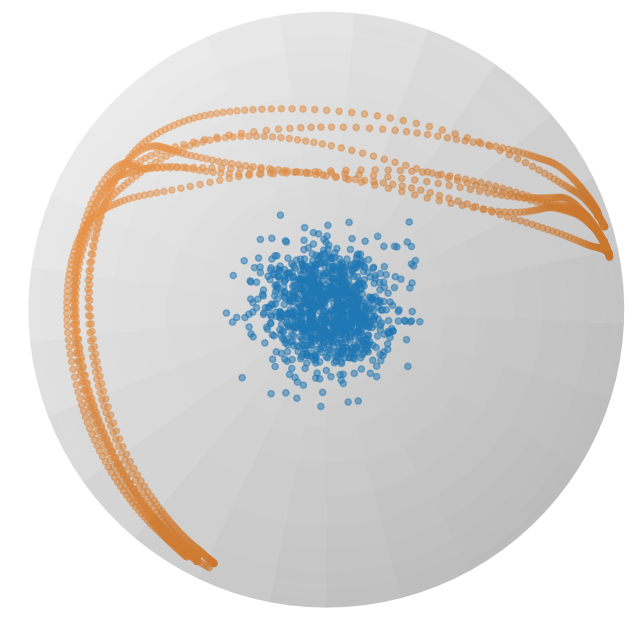} 
\includegraphics[width=0.32\linewidth]{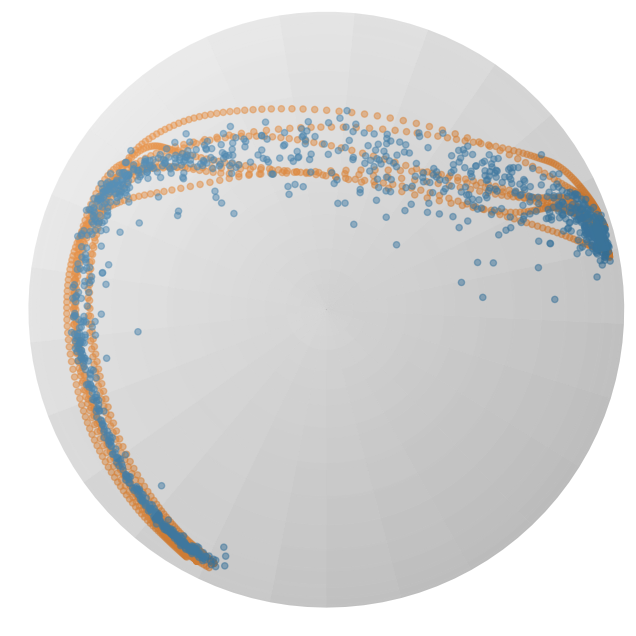}
\includegraphics[width=0.32\linewidth]{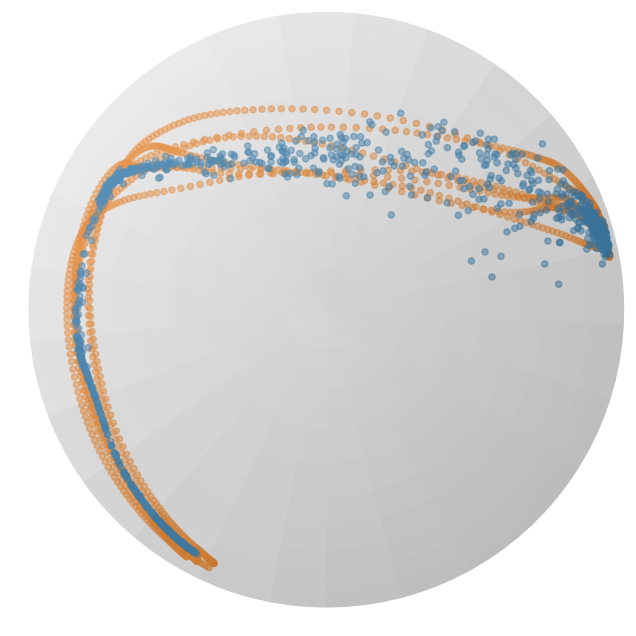}
\caption{Flows of the RFM (\textbf{top}) and SRFM (\textbf{bottom}) trained on the $\mathsf{L}$-shape LASA dataset projected on the sphere manifold. Orange points represent the training dataset, while blue points are sampled from the generated probability path at different times $t=\{0.0, 1.0, 1.5\}$ across the three columns. }
\label{fig:lasaRfmSrfm}
\end{figure}

Although RFM is theoretically designed to match a time-dependent vector field defined over the time horizon $t \in [0, 1]$, we observed that, in practice, its inference performance often improves when evaluating trajectories slightly beyond $t=1$. To investigate this phenomenon, we conduct experiments on two tasks: The Euclidean \textsc{Push-T} task from~\cite{diffpol} and the \textsc{Square} task from the Robomimic robotic manipulation benchmark~\cite{robomimic}. As shown in Figure~\ref{figure:rfmp_beyond_t1}, RFMP performance improves when the ODE integration horizon slightly exceeds $t=1$, but it worsens and eventually gets unstable when integrating for longer time horizons.
This observation motivates the development of SRFMP introduced next, which enhances RFMP by explicitly enforcing inference stability beyond the unit interval using the LaSalle’s invariance principle. This approach stabilizes the policy inference, making it robust to the ODE integration horizon, as illustrated in Figures~\ref{fig:RFMP_overview},~\ref{fig:lasaRfmSrfm} and~\ref{fig:SpdlasaRfmSrfm}.\looseness=-1 

\subsection{Stable Riemannian Flow Matching Policy}
\label{subsec:SRFMP}
Both CFM and RCFM train and integrate the learned vector field $v_t(\vx|\vtheta)$ within the interval $t=[0,1]$. 
However, they do not guarantee that the flow converges stably to the target distribution at $t=1$. Besides, the associated vector field may even display strongly diverging behaviors when going beyond this upper boundary~\cite{stableflow}. The aforementioned issues may arise due to numerical inaccuracies when training or integrating the vector field.
To solve this problem, Sprague \emph{et al.}~\cite{stableflow} proposed Stable Autonomous Flow Matching (SFM), which equips the dynamics of FM with stability to the support of the target distribution. 
Here, we propose to improve RFMP with SFM, which we generalize to the Riemannian case, in order to guarantee that the flow stabilizes to the target policy at $t=1$. Our experiments show that this approach not only enhances RFMP's robustness but also further reduces inference time.\looseness=-1 

\subsubsection{Stable Euclidean FMP}
We first summarize the Euclidean SFM~\cite{stableflow}, and then show how SFM can be integrated into RFMP. 
SFM leverages the stochastic LaSalle's invariance principle~\cite{lasalleinvar, stochasticlasalle} --- a criterion from control theory used to characterize the asymptotic stability of stochastic autonomous dynamical systems --- to design a stable vector field $u$. Sprague \emph{et al.}~\cite[Thm 3.8]{stableflow} adapts this principle to the FM setting as follows. \looseness-1
\begin{theorem}
\label{theory:lasalle}
\textbf{(Stochastic LaSalle's Invariance Principle)} If there exists a time-independent vector field $u$, a flow $\psi$ generated by $u$, and a positive scalar function $H$ such that, 
\begin{equation}
    \mathcal{L}_{\vu} H(\vx)=\nabla_{\vx} H(\vx)u(\vx)\le0,
    \label{equation:LaSalle}
\end{equation}
where $\mathcal{L}_{\vu} H(\vx)$ is the directional derivative of the scalar function $H$ in the direction $\vu$ and $\nabla_{\vx} H$ is the gradient of the function $H$, then, 
\begin{equation*}
    \lim\limits_{t \to \infty}\psi(\vx, t) \in \{ \vx \in \mathcal{X} | \mathcal{L}_u H(\vx) = 0 \},
\end{equation*}
almost surely with $\vx\sim p(\vx, 0)$.
\end{theorem}
Intuitively, Theorem~\ref{theory:lasalle} provides conditions for convergence to an invariant set even when $\mathcal{L}_u H(\vx)$ is not strictly negative, therefore accounting for stochastic fluctuations in the system.
Theorem~\ref{theory:lasalle} notably holds if $u(\vx)$ is a gradient field of $H$, i.e., if $u(\vx) = -\nabla_{\vx} H(\vx)^\top$. In this case, the problem of finding a stable vector field boils down to defining an appropriate scalar function $H$. 
As LaSalle's invariance principle requires an autonomous, a.k.a time-independent, vector field, Sprague \emph{et al.}~\cite{stableflow} augment the FM state space $\vx$ with an additional dimension $\tau$, called temperature or pseudo time, so that the SFM augmented state space becomes $\bm{\xi}=[\vx, \tau]$. 
The pair $(H, \bm{\xi})$ is then defined so that it satisfies~\eqref{equation:LaSalle} as, 
\begin{equation}
    H(\bm{\xi}|\bm{\xi}_1) = \frac{1}{2}(\bm{\xi}-\bm{\xi}_1)^\top \mA(\bm{\xi}-\bm{\xi}_1),
\end{equation}
\begin{equation}
\label{equation:srfm vector field}
    u(\bm{\xi}|\bm{\xi}_1) = -\nabla_{\bm{\xi}} H(\bm{\xi}|\bm{\xi}_1)^\top = -\mA(\bm{\xi}-\bm{\xi}_1),
\end{equation}
where $\mA$ is a positive-define matrix.
To simplify the calculation, $\mA$ is set as the diagonal matrix,
\begin{equation}
    \mA = \left[
        \begin{matrix}
         \lambda_{\vx} \mI & \bm{0} \\
         \bm{0} & \lambda_{\tau}
        \end{matrix}
    \right],
    \label{equation:SRFM_A}
\end{equation}
with $\lambda_{\vx}, \lambda_{\tau} \in \mathbb{R}$. The vector field $u$ and corresponding stable flow $\psi_t$ are then given as,
\begin{equation}
    \label{equation:stablevfeuc}
    u(\bm{\xi}_t|\bm{\xi}_1)= \left[
        \begin{aligned}
            u_{\vx}(\vx_t|\vx_1) \\
         u_\tau(\tau_t|\tau_1)
        \end{aligned}
    \right] = \left[
        \begin{aligned}
            -\lambda_\vx(\vx_t-\vx_1) \\
         -\lambda_\tau(\tau_t-\tau_1)
        \end{aligned}
    \right],
\end{equation}
\begin{equation}
    \label{equation:stable flow}
    \psi_t(\bm{\xi}_0|\bm{\xi}_1) =\left[
    \begin{aligned}
        \psi_t(\vx_0|\vx_1) \\
       \psi_t(\tau_0|\tau_1)
    \end{aligned}
    \right] =\left[
    \begin{aligned}
        \vx_1 + e^{-\lambda_\vx t}(\vx_0 -\vx_1) \\
        \tau_1 + e^{-\lambda_\tau t}(\tau_0 - \tau_1)
    \end{aligned}
    \right].
\end{equation}
The parameters $\tau_1$ and $\tau_0$ define the range of the $\tau$ flow, while the parameters $\lambda_\vx$ and $\lambda_\tau$ determine its convergence. Specifically, the flow converges faster for higher values of $\lambda_{\vx}$ and $\lambda_\tau$. Moreover, the ratio between $\lambda_{\vx}/\lambda_\tau$ determines the relative rate of convergence of the spatial and pseudo-time parts of the flow. We ablate the influence of $\lambda_{\vx}$ and $\lambda_\tau$ on SRFMP in Section~\ref{subsec:pushT}. 

We integrate SFM to RFMP by regressing an observation-conditioned vector field $v(\bm{\xi}|\vo; \vtheta)$ to a stable target vector field $u(\bm{\xi}_t| \bm{\xi}_1)$, where $\bm{\xi} = [\va^s,...,\va^{s+T_p}, \tau]$ is the augmented prediction horizon vector, and $\vo$ is the observation vector.

\begin{figure}[t]
    \centering
    \includegraphics[trim={20cm 7.0cm 40.0cm 8.0cm},clip,width=0.32\linewidth]{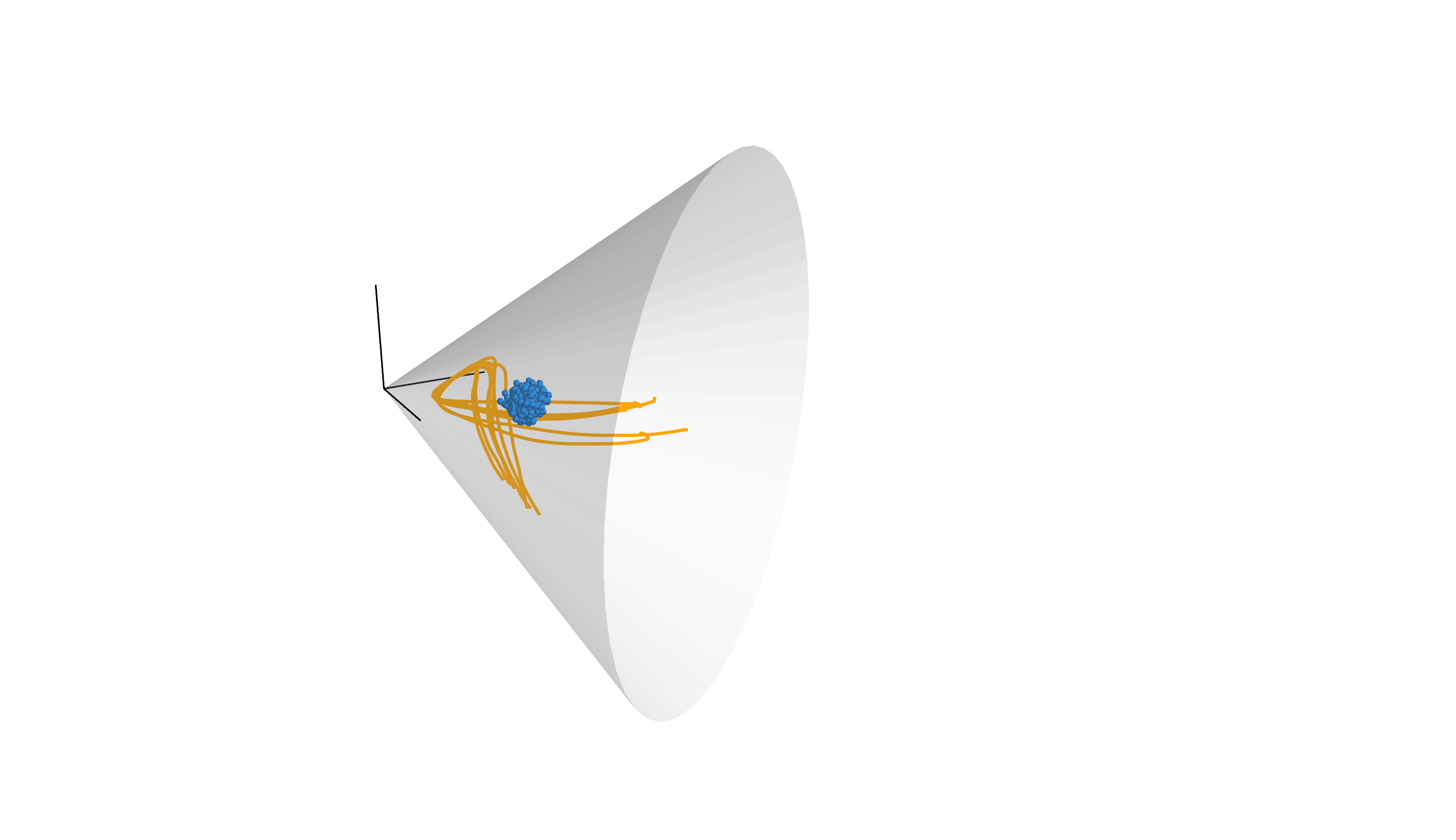} 
    \includegraphics[trim={20cm 7.0cm 40.0cm 8.0cm},clip,width=0.32\linewidth]{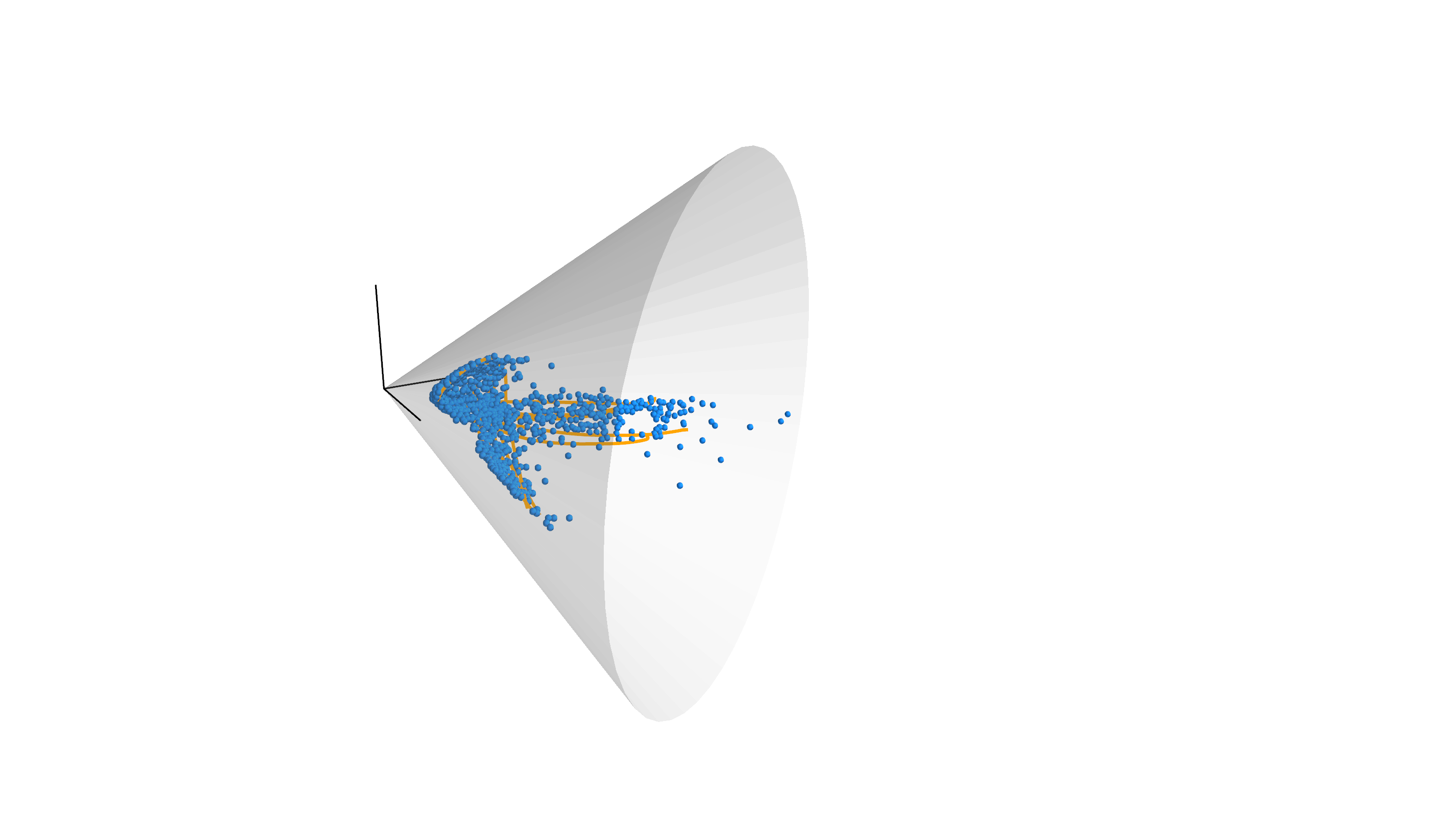}
    \includegraphics[trim={15cm 5.0cm 45.0cm 10.0cm},clip,width=0.32\linewidth]{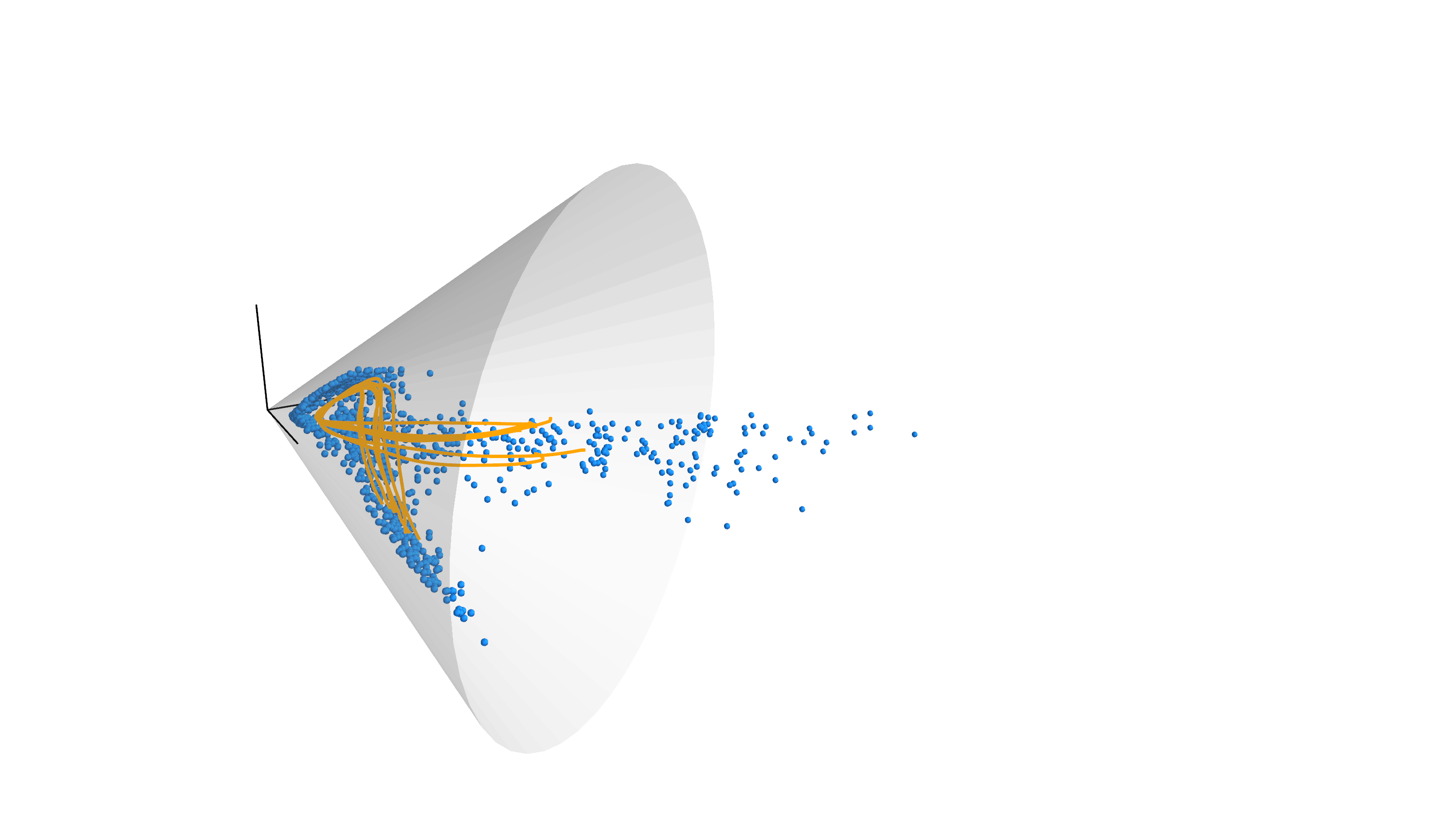} \\
    \includegraphics[trim={20cm 8.0cm 40.0cm 7.0cm},clip,width=0.32\linewidth]{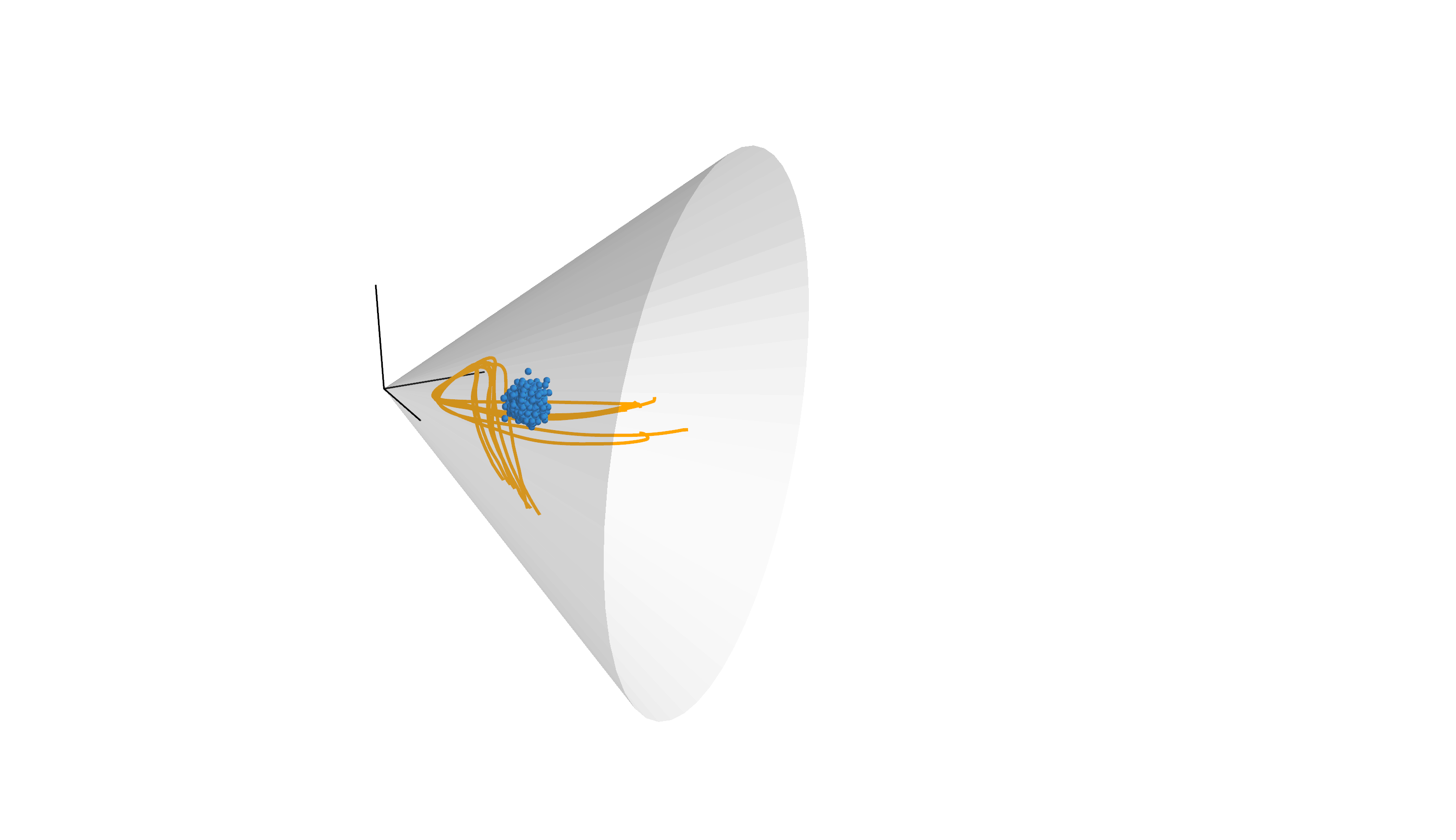} 
    \includegraphics[trim={20cm 8.0cm 40.0cm 7.0cm},clip,width=0.32\linewidth]{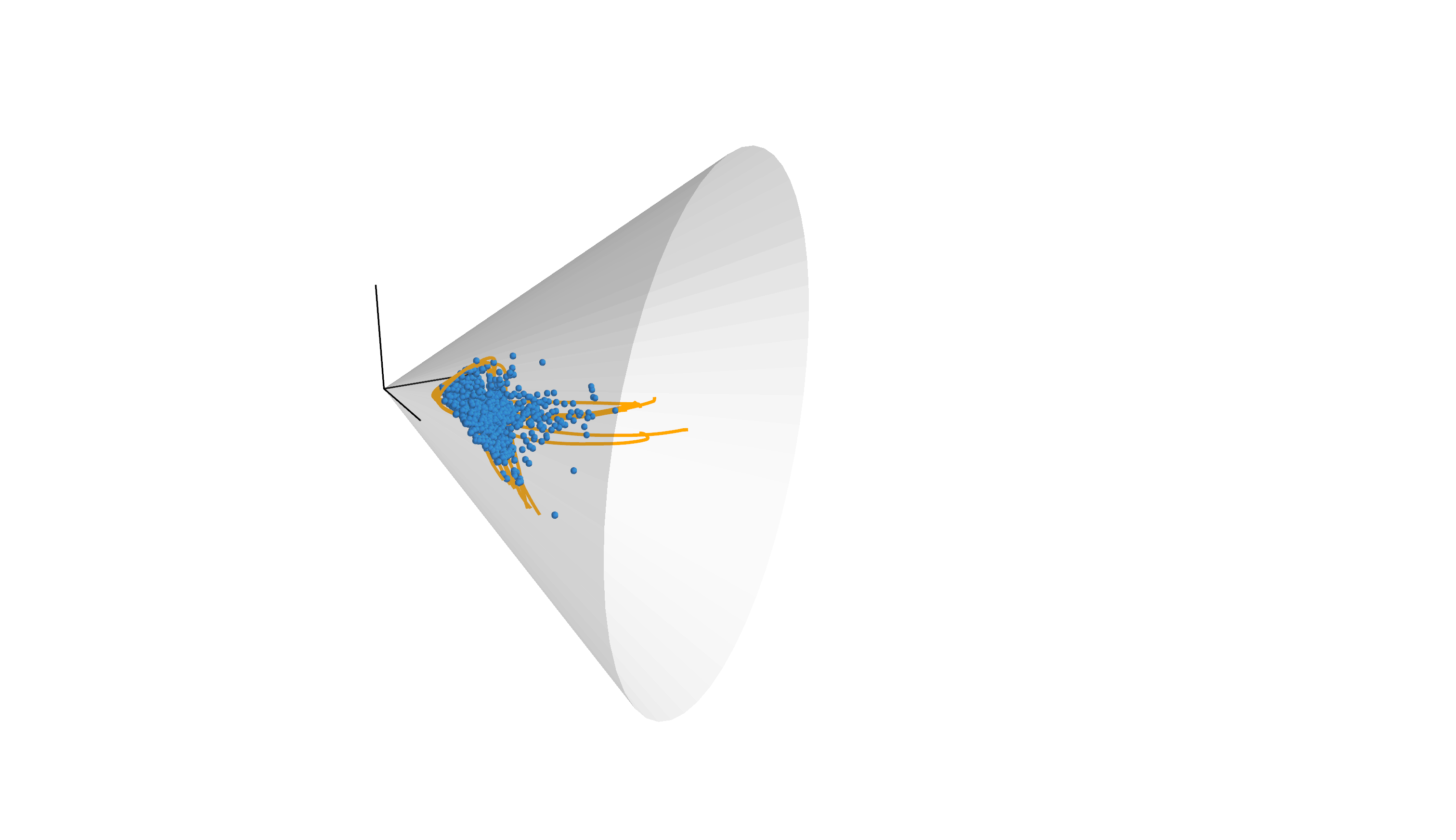}
    \includegraphics[trim={20cm 8.0cm 40.0cm 7.0cm},clip,width=0.32\linewidth]{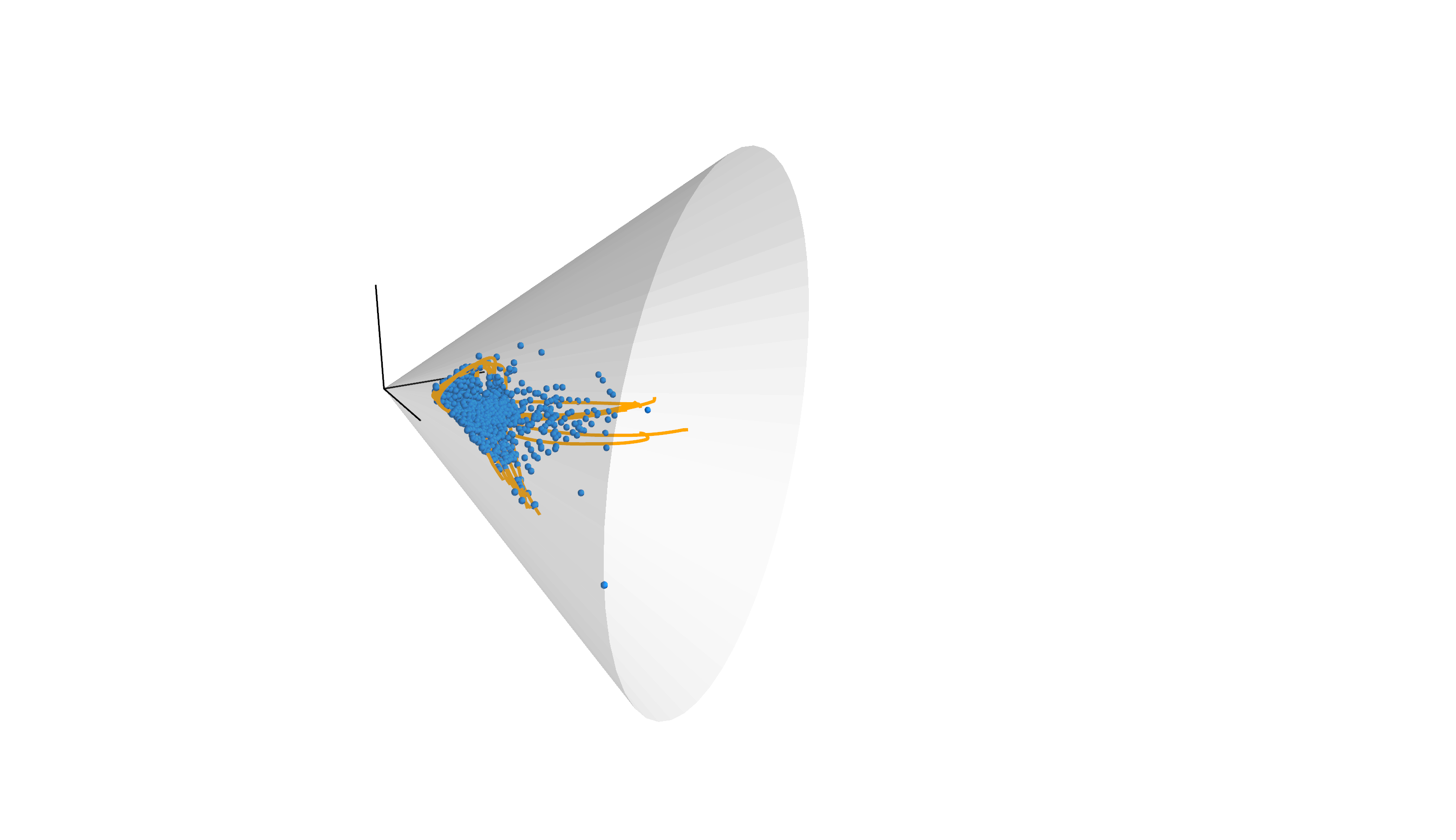}
    \caption{Flows of the RFM (\textbf{top}) and SRFM (\textbf{bottom}) on the SPD manifold $\mathcal{S}_{++}^2$. Orange points represent the training dataset, while blue points correspond to sampled from the generated probability path at different times $t=\{0.0, 1.0, 1.5\}$ across the three columns.}
    \label{fig:SpdlasaRfmSrfm}
\end{figure}

\subsubsection{Stable Riemannian FMP}
\begin{algorithm}[tb]
    \label{algorithm:srfmp training}
    \caption{SRFMP Training \& Inference}
    \textbf{Training} \\
    \KwIn{Initial parameters $\vtheta$, prior and target distributions $p_0$, $p_1$.}
    \KwOut{Learned vector field parameters $\vtheta$.} 
    \While{termination condition unsatisfied}{
        Sample flow time step $t$ from an uniform distribution $\mathcal{U}[0 ,1]$. \\
        Sample noise $\va_0 \sim p_0$. \\
        Jointly sample action sequence $\va_1\sim p_1$ and corresponding observation vector $\vo$. \\
        Form the vectors $\bm{\xi}_1=[\va_1, \tau_1]$ and $\bm{\xi}_0=[\va_0, \tau_0]$.\\
        Compute conditional vector field $u_t(\bm{\xi}_t|\bm{\xi}_1)$ via~\eqref{equation:srfm vector field}.\\
        Evaluate the loss $\ell_{\text{SRFMP}}$ as defined in~\eqref{eq:SRFMP_loss}.\\
        Update parameters $\vtheta$.
        }
    \textbf{Inference Step}\\
    \KwIn{Predefined number of function evaluation $N$, learned vector field $v_{\vtheta}$, observation vector $\vo$, prior distribution $p_0$.}
    Sample $\va_0 \sim p_0$, and set $k=1$, $t=0$, $\bm{\xi}_0=[\va_0, \tau_0]$.\\
    \While{$k \leq N$}{
        
        \If{$k=1$}{
            $\Delta t = \frac{1}{\lambda_{\vx}}$
        }
        \Else{
            $\Delta t = \varepsilon \leq \frac{1}{\lambda_{\vx}}$ 
        }
        Integrate the learned Riemannian vector field $\bm{\xi}_{t + \Delta t} = \text{Exp}_{\bm{\xi}_{t}} (v_{\vtheta}(\bm{\xi}_{t}, \vo) \Delta t)$. \\
        Update time $t = t + \Delta t$. \\ 
        Update iteration $k = k + 1$.
    }
\end{algorithm}
Next, we introduce our extension, stable Riemannian flow matching (SRFM), which generalizes SFM~\cite{stableflow} to Riemannian manifolds. 
Similarly as SFM, we define a time-invariant vector field $u$ by augmenting the state space with the pseudo-time state $\tau$, so that $\bm{\xi}=[\vx, \tau]$. Notice that, in this case, $\vx\in\manifold$ and thus $\bm{\xi}$ lies on the product of Riemannian manifolds $\manifold\times\euclideanspace$. 
Importantly, Theorem~\ref{theory:lasalle} also holds for Riemannian autonomous systems, in which case $\nabla_{\vx} H(\vx)$ denotes the Riemannian gradient of the positive scalar function $H$. We formulate $H$ so that the pair $(H,\bm{\xi})$ satisfies~\eqref{equation:LaSalle} as,
\begin{equation}
    H(\bm{\xi}|\bm{\xi}_1) = \frac{1}{2} \logmap{\bm{\xi}_1}{\bm{\xi}}^\top \mA \logmap{\bm{\xi}_1}{\bm{\xi}},
\end{equation}
which leads to the Riemannian vector field,
\begin{equation}
     u(\bm{\xi}|\bm{\xi}_1) = - \nabla_{\bm{\xi}}H(\bm{\xi}|\bm{\xi}_1)^\top =  -\mA \logmap{\bm{\xi}_1}{\bm{\xi}}.
     \label{eq:RiemannianVectorField}
\end{equation}
By setting the positive-definite matrix $\mA$ as in~\eqref{equation:SRFM_A}, we obtain the Riemannian vector field,
\begin{equation}
    \label{equation:stableriemanianvectorfield}
    u(\bm{\xi}_t|\bm{\xi}_1)= \left[
        \begin{aligned}
            u_\vx(\vx_t|\vx_1) \\
         u_\tau(\tau_t|\tau_1)
        \end{aligned}
    \right] = \left[
        \begin{aligned}
         -\lambda_\vx \logmap{\vx_1}{\vx_t} \\
         -\lambda_\tau(\tau_t -\tau_1)
        \end{aligned}
    \right],
\end{equation}
which generates the stable Riemannian flow,
\begin{equation}
    \label{equation:stableriemanianflow}
    \psi_t(\bm{\xi}_0|\bm{\xi}_1) =\left[
    \begin{aligned}
        \psi_t(\vx_0|\vx_1) \\
       \psi_t(\tau_0|\tau_1)
    \end{aligned}
    \right]= \left[
    \begin{aligned}
       \text{Exp}_{\vx_1}\left(e^{-\lambda_\vx t}\text{Log}_{\vx_1}(\vx_0) \right)\\
        \tau_1 + e^{-\lambda_\tau t}(\tau_0 - \tau_1)
    \end{aligned}
    \right].
\end{equation}
The parameters $\lambda_\vx$ and $\lambda_\tau$ have the same influence as in the Euclidean case.
Notice that the spatial part of the stable flow~\eqref{equation:stableriemanianflow} closely resembles the geodesic flow~\eqref{equation:rfmflow} proposed in~\cite{riemannianfm}. 
Figures~\ref{fig:lasaRfmSrfm} and~\ref{fig:SpdlasaRfmSrfm} show examples of learned RFM and SRFM flows at times $t=\{0, 1, 1.5\}$ on the sphere and symmetric positive-definite (SPD) matrices manifold. The RFM flow diverges from the target distribution at time $t>1$, while the SRFM flow is stable and adheres to the target distribution for $t \geq 1$. 

Finally, the process to induce this stable behavior into the RFMP flow involves two main changes: \emph{(1)} We define an augmented action horizon vector $\bm{\xi} = \left[\va^s,\ldots,\va^{s+T_p}, \tau\right]$, and; \emph{(2)} We regress the observation-conditioned Riemannian vector field $v(\bm{\xi}|\vo; \vtheta)$ against the stable Riemannian vector field $u(\bm{\xi}|\bm{\xi}_1)$ defined in~\eqref{eq:RiemannianVectorField}, where $\vo$ denotes the observation vector. 
The model is then trained to minimize the SRFMP loss,
\begin{equation}
     \ell_{\text{SRFMP}} = \mathbb{E}_{t, q(\va_1), p(\va_0)}\Arrowvert v(\bm{\xi}_t|\vo; \vtheta)-u(\bm{\xi}_t|\bm{\xi}_1)\Arrowvert_{g_{\va_t}}^{2} .
    \label{eq:SRFMP_loss}
\end{equation}
This approach, hereinafter referred to as stable Riemannian flow matching policy (SRFMP), is summarized in Algorithm~\ref{algorithm:srfmp training}. 
The learned SRFMP vector field drives the flow to converge to the target distribution within a certain time horizon, ensuring that it remains within this distribution, as illustrated by the bottom row of Figures~\ref{fig:RFMP_overview},~\ref{fig:lasaRfmSrfm}, and~\ref{fig:SpdlasaRfmSrfm} . In contrast, the RFMP vector field may drift the flow away from the target distribution at $t>1$ (see the top row of Figures~\ref{fig:RFMP_overview},~\ref{fig:lasaRfmSrfm}, and~\ref{fig:SpdlasaRfmSrfm}). Therefore, SRFMPs provide flexibility and increased robustness in designing the generation process, while RFMPs are more sensitive to the integration process.

\subsubsection{Solving the SRFMP ODE}
As previously discussed, querying SRFMP policies involves integrating the learned vector field along the time interval $t=[0,T]$ with time boundary $T$. To do so, we use the projected Euler method, which integrates the vector field on the tangent space for one Euler step and then projects the resulting vector onto the manifold $\manifold$. Assuming an Euclidean setting and that $v(\bm{\xi}|\vo; \vtheta)$ is perfectly learned, this corresponds to recursively applying,\looseness=-1  
\begin{equation}
\label{eq:SFMrecursion}
    \vx_{t+1} = \vx_t + v_{\vx}(\vx_t|\vo; \vtheta) \Delta t \, \approx \, \vx_t + \lambda_\vx (\vx_1 - \vx_t) \Delta t , 
\end{equation}
with time step $\Delta t$ and $v_{\vx}$ following the same partitioning as $u_{\vx}$ in~\eqref{equation:stableriemanianvectorfield}. The time step $\Delta t$ is typically set as $\Delta t=T/N$, where $N$ is the total number of ODE steps. 
Here we propose to leverage the structure of SRFMP to choose the time step $\Delta t$ in order to further speed up the inference time of RFMP.
Specifically, we observe that the recursion~\eqref{eq:SFMrecursion} leads to,
\begin{equation}
\label{eq:SFMrecursionNsteps}
   \vx_t = (1-\lambda_{\vx} \Delta t)^{n}(\vx_0-\vx_1) + \vx_1,
\end{equation}
after $n$ time steps. It is easy to see that $\vx_t$ converges to $\vx_1$ after a single time step when setting $\Delta t=1/ \lambda_{\vx}$. 

In the Riemannian case, assuming that $v(\bm{\xi}|\vo; \vtheta)$ approximately equals the Riemannian vector field~\eqref{equation:stableriemanianvectorfield}, we obtain,
\begin{equation}
    \vx_{t+1} = \text{Exp}_{\vx_t}\left(v_{\vx}(\vx|\vo; \vtheta) \Delta t\right) \, \approx \, \text{Exp}_{\vx_t}\left(\lambda_\vx \text{Log}_{\vx_t}(\vx_1) \Delta t\right) .
\end{equation}
Similarly, it is easy to see that the Riemannian flow converges to $\vx_1$ after a single time step for $\Delta t=1/ \lambda_{\vx}$.
Importantly, this strategy assumes that the learned vector field is perfectly learned and thus equals the target vector field. However, this is often not the case in practice. However, our experiments show that the flow obtained solving the SRFMP ODE with a single time step $\Delta t=1/ \lambda_\vx$ generally leads to the target distribution. In practice, we set $\Delta t$ to $1/ \lambda_\vx$ for the first time step, and to a smaller value afterwards for refining the flow. 

\section{Experiments}
\label{sec:experiments}
We thoroughly evaluate the performance of RFMP and SRFMP on a set of eight simulation settings and two real-world tasks. The simulated benchmarks are: \emph{(1)} The \textsc{Push-T} task from~\cite{diffpol}; \emph{(2)} A \textsc{Sphere Push-T} task, which we introduce as a Riemannian benchmark; \emph{(3)}-\emph{(7)} Five tasks (\textsc{Lift}, \textsc{Can}, \textsc{Square}, \textsc{Tool Hang}, and \textsc{Transport}) from the large-scale robot manipulation benchmark Robomimic~\cite{robomimic}; and  \emph{(8)} the \textsc{Franka Kitchen} benchmark~\cite{d4rl} featuring complex, long-horizon tasks. The real-world robot tasks correspond to: \emph{(1)} A \textsc{Pick \& Place} task; and (2) A \textsc{Mug Flipping} task. Collectively, these ten tasks serve as a benchmark to evaluate \emph{(1)} the performance, \emph{(2)} the training time, and \emph{(3)} the inference time of RFMP and SRFMP with respect to state-of-the-art generative policies, i.e., DP and extensions thereof.

\begin{table*}[tbp]
  \centering
  \caption{Hyperparameters for all experiments: Resolution of original and cropped image, number of parameters for the learned vector field (VF) using RFMP and SRFMP, number of ResNet parameters, training epochs, and batch size.}
  \label{table:hyperParamRFMP}
  \footnotesize
  \resizebox{\linewidth}{!}{ % Consider \textwidth for better width distribution
    \begin{tabular}{l c c c c c c c}
      %\toprule
      \rowcolor{gray!15} % Light gray for the header row
      \textbf{Experiment} &  \textbf{Image res.} & \textbf{Crop res.} & \textbf{RFMP VF \# params} & \textbf{SRFMP VF \# params} & \textbf{ResNet \# params} & \textbf{Epochs} & \textbf{Batch size}\\ 
      \rowcolor{gray!15} % Light gray for the header row
      \multicolumn{8}{l}{\textbf{\textsc{Push-T} Tasks}} \\
      Euclidean \textsc{Push-T}         &  $96 \times 96$  &  $84 \times 84$ &  $\num{8.0e+07}$  &  \num{8.14e+07}& $\num{1.12e+07}$  & $300$ & $256$  \\
      \rowcolor{gray!5} % Light gray for alternating rows
      \textsc{Sphere Push-T}     &  $100 \times 100$  &  $84 \times 84$ &  $\num{8.0e+07}$  &  \num{8.14e+07}&$\num{1.12e+07}$ & $300$ & $256$ \\
      \rowcolor{gray!15} % Light gray for the header row
      \multicolumn{8}{l}{\textbf{State-based simulation tasks}} \\
      \rowcolor{gray!5} % Light gray for alternating rows
      Robomimic \textsc{Lift}         & N.A. &  N.A. & $\num{6.58e+07}$ & \num{6.68e+07} &N.A.& 50 & 256 \\
      Robomimic \textsc{Can}           & N.A.  &  N.A.  & $\num{6.58e+07}$ &\num{6.68e+07} & N.A. & $50$  & $256$\\
      \rowcolor{gray!5} % Light gray for alternating rows
      Robomimic \textsc{Square}         & N.A. &  N.A. & $\num{6.58e+07}$ &\num{6.68e+07} & N.A.  & $50$ & $512$ \\
      Robomimic \textsc{Tool Hang}     & N.A. & N.A. & $\num{6.58e+07}$&\num{6.68e+07}  & N.A.& $100$ & $512$ \\
      \rowcolor{gray!5} % Light gray for alternating rows
      \textsc{Franka  Kitchen} & N.A. & N.A. & \num{6.69e+07} & \num{6.89e+07}& N.A. & 500 & 128
       \\
      \rowcolor{gray!15} % Light gray for the header row
      \multicolumn{8}{l}{\textbf{Vision-based simulation tasks}} \\
      Robomimic \textsc{Lift}  & $2 \times 84 \times 84$   & $2 \times 76 \times 76$ & $\num{9.48e+07}$& \num{9.69e+07} &$2\times \num{1.12e+07}$& 100 & 256
      \\
      \rowcolor{gray!5} 
      Robomimic \textsc{Can}   & $2 \times 84 \times 84$& $2 \times 76 \times 76$ &$\num{9.48e+07}$ & \num{9.69e+07} &$2\times \num{1.12e+07}$& 100 & 256 \\
      Robomimic \textsc{Square}   &$2 \times 84 \times 84$ &  $2 \times 76 \times 76$& $\num{9.48e+07}$& \num{9.69e+07} &$2\times \num{1.12e+07}$& 100 & 512
      \\
      \rowcolor{gray!5} 
      Robomimic \textsc{Transport}  & $4 \times 84 \times 84$ & $4 \times 76 \times 76$ & $\num{1.24e+08}$ & \num{1.27e+08} & $4\times \num{1.12e+07}$& 200 & 256 \\
      \rowcolor{gray!15} % Light gray for the header row
      \multicolumn{8}{l}{\textbf{Real-world experiments}} \\
      \rowcolor{gray!5} 
      \textsc{Pick \& place}  &  $320 \times 240$  &  $288 \times 216$ &  $\num{2.51e+07}$ &\num{2.66e+07} &  $\num{1.12e+07}$  & $300$ & $256$  \\
      \textsc{Mug flipping}        &  $320 \times 240$  & $256 \times 192$ &  $\num{2.51e+07}$ & \num{2.66e+07}&  $\num{1.12e+07}$   & $300$ & $256$ \\

      %\bottomrule
    \end{tabular}
  }
\end{table*}

\subsection{Implementation Details}
To establish a consistent experimental framework, we first introduce the neural network architectures employed in RFMP and SRFMP across all tasks. We then describe the considered baselines, and our overall evaluation methodology.

\subsubsection{RFMP and SRFMP Implementation}
\label{subsubsec:sRFMPimplementation}
Our RFMP implementation builds on the RFM framework from Chen and Lipman~\cite{riemannianfm}. 
We parameterize the vector field $v_t(\bm{a}|\bm{o}; \bm{\theta})$ using the UNet architecture employed in DP~\cite{diffpol}, which consists of $3$ layers with downsampling dimensions of $(256, 512, 1024)$ and $(128, 256, 512)$ for simulated and real-world tasks. Each layer employs a $1$-dimensional convolutional residual network as proposed in~\cite{janner2022planningdiffusion}. We implement a Feature-wise Linear Modulation (FiLM)~\cite{perez2018film} to incorporate the observation condition vector $\vo$ and time step $t$ into the UNet. Instead of directly feeding the FM time step $t$ as a conditional variable, we first project it into a higher-dimensional space using a sinusoidal embedding module, similarly to DP. 
For tasks with image-based observations, we leverage the same vision perception backbone as in DP~\cite{diffpol}. 
Namely, we use a standard \mbox{ResNet-$18$} in which we replace: \emph{(1)} the global average pooling with a spatial softmax pooling, and \emph{(2)} BatchNorm with GroupNorm. 
Our SRFMP implementation builds on the SFM framework~\cite{stableflow}. We implement the same UNet as RFMP to represent $v_{\vx}$ by replacing the time step $t$ by the temperature parameter $\tau$. We introduce an additional Multi-Layer Perceptron (MLP) to learn $v_\tau$. As for $t$ in RFMP, we employed a sinusoidal embedding for the input $\tau$. The boundaries $\tau_0$ and $\tau_1$ are set to $0$ and $1$ in all experiments.

\begin{figure}[tbp]
\centering
\includegraphics[width=\linewidth,trim={0.0cm 1.5cm 0.0cm 0.0cm},clip]{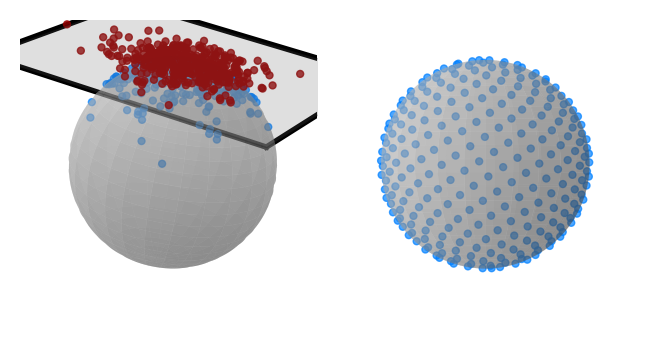}
\caption{2D visualization of prior distributions on the sphere: wrapped Gaussian distribution (\textbf{left}) and sphere uniform distribution (\textbf{right}). The sphere Gaussian distribution is obtained by first sampling from a Euclidean Gaussian distribution on a tangent space of the sphere (the red points), followed by the exponential map, which projects the samples onto the sphere manifold $\mathcal{S}^d$ (blue points). The spherical uniform distribution is computed by normalizing samples from a zero-mean Euclidean Gaussian distribution.}
\label{figure:sphere_prior}
\end{figure}

We implement different prior distributions for different tasks. For the Euclidean \textsc{Push-T} and the Robomimic tasks, the action space is $\euclideanspace^d$ and we thus define the prior distribution as a Euclidean Gaussian distribution for both RFMP and SRFMP. For the \textsc{Sphere Push-T}, the action space is the hypersphere $\mathcal{S}^{2}$. In this case, we test two types of Riemannian prior distribution, namely a spherical uniform distribution, and a wrapped Gaussian distribution~\cite{Mardia99:DirectionalStats, wrappedgaussian}, illustrated in Figure~\ref{figure:sphere_prior}. For the \textsc{Franka kitchen} benchmark, we define the action space as the product of manifolds $\manifold=\euclideanspace^3  \times \mathcal{S}^{3} \times \euclideanspace^2$, whose components represent the end-effector position, the end-effector orientation (encoded as quaternions), and the gripper fingers position. Regarding the real robot tasks, the action space is defined similarly as the product of manifolds $\manifold=\euclideanspace^3  \times \mathcal{S}^{3} \times \euclideanspace^1$, whose components represent the position, orientation (encoded as quaternions), and opening of the gripper. For the \textsc{Franka kitchen} benchmark and the real robot tasks, the Euclidean and hypersphere parts employ Euclidean Gaussian distributions and a wrapped Gaussian distribution, respectively. 
Notice that this choice of prior distributions follows common practice in flow matching literature, where Gaussian and uniform priors are widely adopted due to their simplicity and effectiveness~\cite{riemannianfm}. While more sophisticated approaches could be explored, e.g., using Gaussian Process as prior distributions~\cite{fmgpprior}, or learning task-dependent priors as in D-Flow~\cite{dflow}, the choice depends on the problem at hand. We focus here on standard priors to maintain simplicity and consistency across tasks, while isolating their impact on our framework.\looseness-1

For all experiments, we optimize the network parameters of RFMP and SRFMP using AdamW~\cite{adamw} with a learning rate of $\eta\!=\!1\!\times\! 10^{-4}$ and weight decay of $w_d\!=\!0.001$ based on an exponential moving averaging (EMA) framework on the weights~\cite{ema} with a decay of $w_{\text{EMA}}\!=\!0.999$. We set the SRFMP parameters as $\lambda_{\vx}\!=\!\lambda_{\tau}\!=\!2.5$. Table~\ref{table:hyperParamRFMP} summarizes the image resolution, number of parameters, and number of training epochs used in each experiment. 
Note that the policies predicts a sequence of actions $\bm{a}$ over a given time horizon $T_p$. Therefore, the total action space corresponds to the Cartesian product of manifolds computed over the prediction horizon defined for each task. Table~\ref{table:manifold dimension} reports the total task dimensionality $\dim(\bm{a})$ as a function of the action space dimension $\dim(\bm{a}^s)$ and the prediction horizon $T_p$ for all experiments considered in the paper. Notably, the \textsc{Franka Kitchen} benchmark involves learning a policy that predicts a $1944$-dimensional action sequence, enabling us to test the scalability of our method in high-dimensional, long-horizon settings. 
We use an action horizon $T_a\!=\!T_p/2$, and an observation horizon $T_o\!=\!2$ for all tasks.

\subsubsection{Baselines}
In~\cite{diffpol}, DP is trained using either Denoising Diffusion Probabilistic Model (DDPM)~\cite{ho2020ddpm} or Denoising Diffusion Implicit Model (DDIM)~\cite{song2020ddim}. In this paper, we prioritize faster inference and thus employ DDIM-based DP for all our experiments. 
We train DDIM with $100$ denoising steps. The prior distribution is a standard Gaussian distribution unless explicitly mentioned. During training we use the same noise scheduler as in~\cite{diffpol}, the optimizer AdamW with the same learning rate and weight decay as for RFMP and SRFMP.
Note that DP does not handle data on Riemannian manifolds, and thus does not guarantee that the resulting trajectories lie on the manifold of interest for tasks with Riemannian action spaces, e.g., the \textsc{Sphere Push-T}, the \textsc{Franka kitchen} benchmark, and the real-world robot experiments. In these cases, we post-process the trajectories obtained during inference and project them on the manifold. In the case of the hypersphere manifold, the projection corresponds to a unit-norm normalization. 

We also compare RFMP and SRFMP against CP~\cite{consistencypolicy} on the Robomimic tasks with vision-based observations. 
For a fair comparison, we retrained CP from scratch using the code provided in~\cite{consistencypolicy}. We pretrain the DP teacher policy for $100$ epochs and subsequently distilled the student policy for another $100$ epochs, following the CP training procedure~\cite{consistencypolicy}. Notice that this increases the total training time, effectively doubling the number of epochs, compared to RFMP, SRFMP, and DP. 
The neural network architecture used in DP and CP to model the diffusion noise is identical to the one used in RFMP and SRFMP. This isolates the influence of the learning algorithm, eliminating the architectural aspects as confounding factors.\looseness=-1 

\begin{figure*}[htb]
\centering
\includegraphics[width=0.13\linewidth]{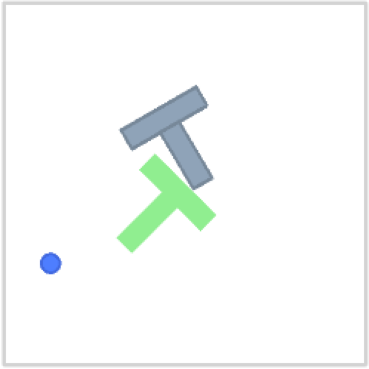}
\hfill
\includegraphics[width=0.13\linewidth]{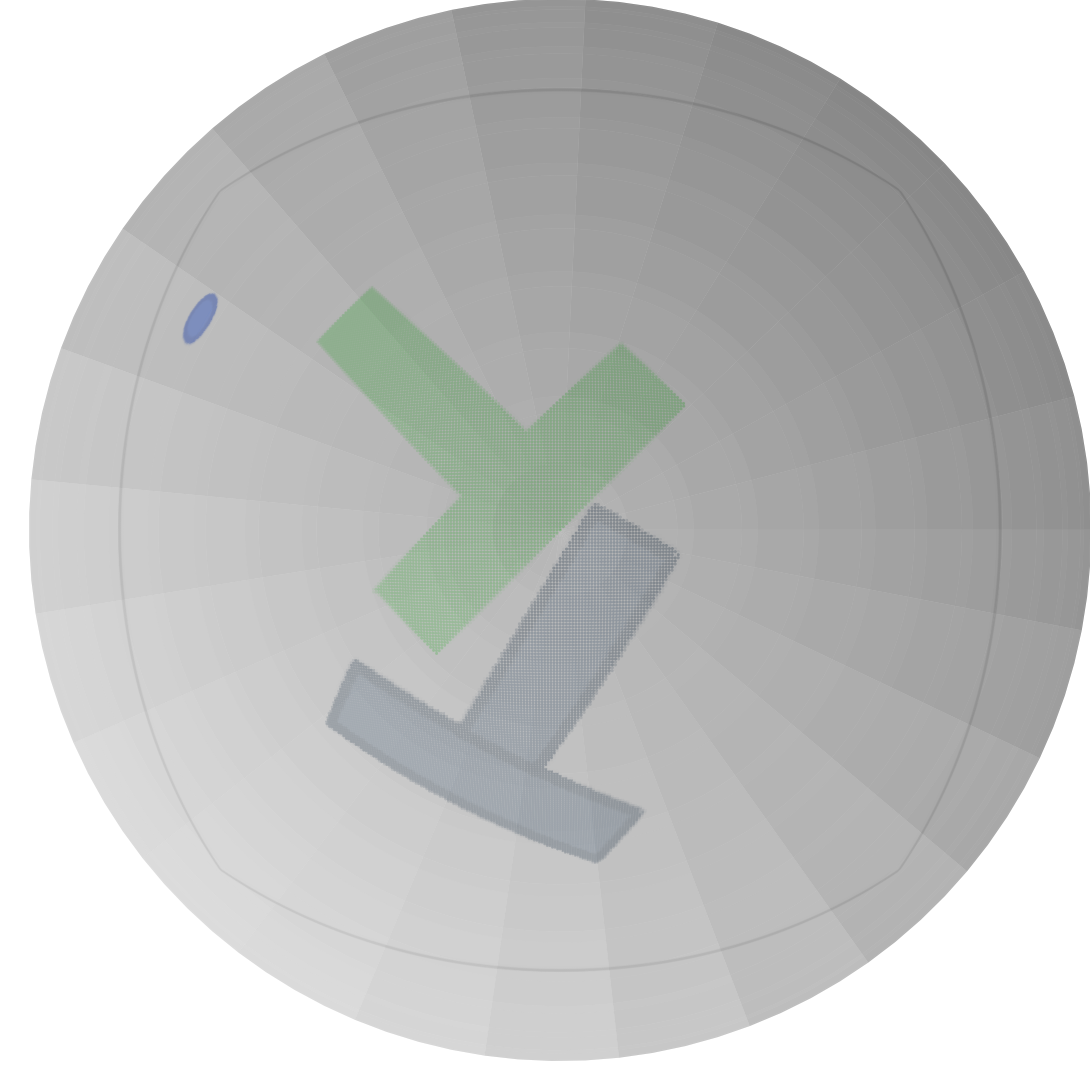}
\hfill
\includegraphics[width=0.13\linewidth]{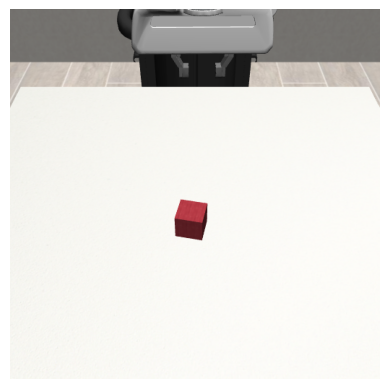}
\hfill
\includegraphics[width=0.13\linewidth]{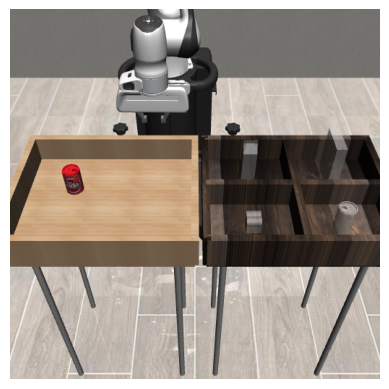}
\hfill
\includegraphics[width=0.13\linewidth]{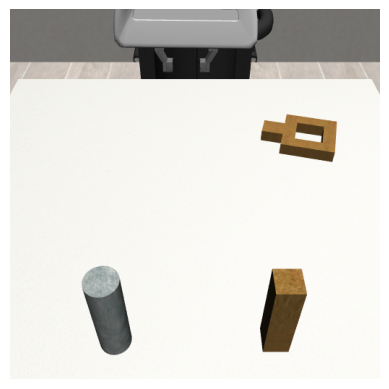}
\hfill
\includegraphics[width=0.13\linewidth]{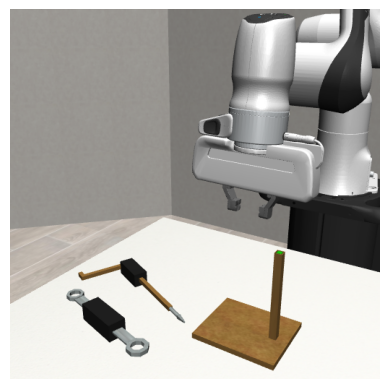}
\hfill
\includegraphics[width=0.13\linewidth]{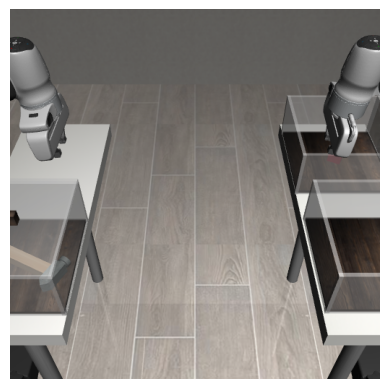}
\vspace{0.2cm}

\includegraphics[width=0.13\linewidth]{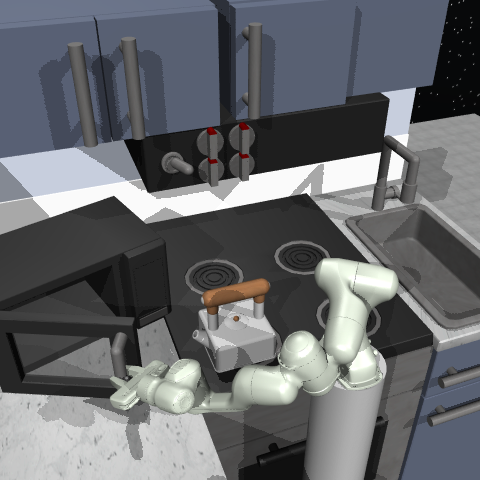}
\hspace{0.2cm}
\includegraphics[width=0.13\linewidth]{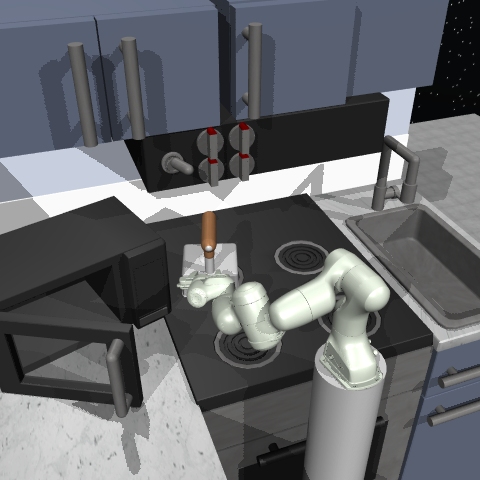}
\hspace{0.2cm}
\includegraphics[width=0.13\linewidth]{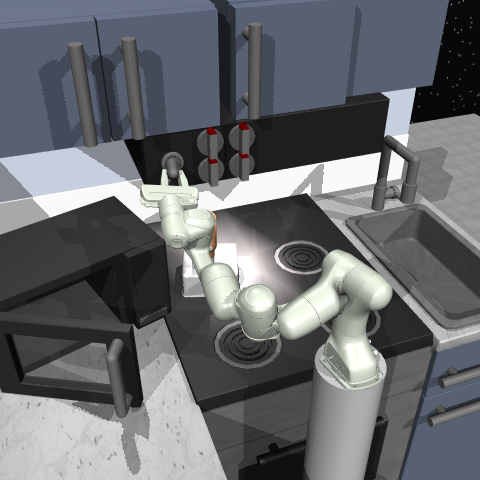}
\hspace{0.2cm}
\includegraphics[width=0.13\linewidth]{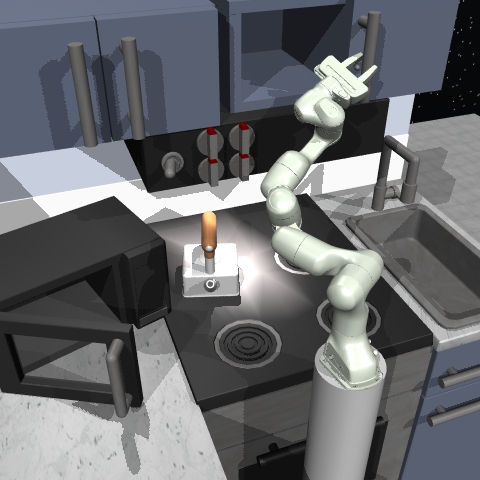}
\caption{Simulation benchmarks. \textbf{Top row}: Euclidean \textsc{Push-T}~\cite{diffpol}, \textsc{Sphere Push-T}, and five Robomimic tasks~\cite{robomimic}: \textsc{Lift}, \textsc{Can}, \textsc{Square}, \textsc{Tool Hang}, \textsc{Transport}. \textbf{Bottom row}: Tasks of the Franka kitchen benchmarks, i.e., open microwave, put kettle on top burner, switch on the light, slide cabinet. }
\label{figure:simulation_tasks}
\end{figure*}

\begin{table}[tbp]
\centering
\caption{Prediction horizon and action space dimensionality. 
}
\label{table:manifold dimension}
  \resizebox{.8\linewidth}{!}{
    \begin{tabular}{lccc}
    \rowcolor{gray!15} \textbf{Task} & \textbf{$T_p$} & $\dim(\bm{a}^s)$  & \textbf{$\dim(\bm{a})$}\\
    Euclidean \textsc{Push-T} & $16$ & $2$ & $32$ \\
    \rowcolor{gray!5} \textsc{Sphere Push-T} & $16$ & $2$ & $32$ \\
    Robomimic tasks & $16$ & $7$ & $112$ \\
    \rowcolor{gray!5} \textsc{Franka Kitchen} & $216$ & $9$ & $1944$ \\
    Real-world experiments & $16$ & $7$ & $112$ \\
    \end{tabular}
}
\end{table}

\subsubsection{Evaluation methodology}
We evaluate all policies using three key metrics: \emph{(1)} The performance, computed as the average task-depending score across all trials, with $50$ trials for each simulated task, and $10$ trials for each real-world task; \emph{(2)} The number of training epochs; and \emph{(3)} The inference time. 
To provide a consistent measure of inference time across RFMP, SRFMP, and DP, we report it in terms of the number of function evaluations (NFE), which is proportional to the inference process time. Given that each function evaluation takes approximately the same time across all methods, inference time comparisons can be made directly based on NFE. For example, in real-world tasks, with an NFE of $2$, DP requires around $0.0075\si{\second}$, while RFMP and SRFMP take approximately $0.0108\si{\second}$ and $0.0113\si{\second}$, respectively. When NFE is increased to $5$, DP takes around $0.0175\si{\second}$, while RFMP and SRFMP require about $0.021\si{\second}$ and $0.0212\si{\second}$.

\subsection{Push-T Tasks}
\label{subsec:pushT}
We first consider two simple \textsc{Push-T} tasks, namely the Euclidean \textsc{Push-T} proposed in~\cite{diffpol}, which was adapted from the Block Pushing task~\cite{florence2022implicit}, and the \textsc{Sphere Push-T task}, which we introduce shortly. 
The goal of the Euclidean \mbox{\textsc{Push-T}} task, illustrated in Figure~\ref{figure:simulation_tasks}, is to push a gray T-shaped object to the designated green target area with a blue circular agent. The agent's movement is constrained by a light gray square boundary. Each observation $\vo$ is composed of the $96\times 96$ RGB image of the current scene and the agent's state information. 
We introduce the \textsc{Sphere Push-T} task, visualized in Figure~\ref{figure:simulation_tasks}, to evaluate the performance of our models on the sphere manifold. Its environment is obtained by projecting the Euclidean \textsc{Push-T} environment on one half of a $2$-dimensional sphere $\mathcal{S}^2$ of radius of $1$. This is achieved by projecting the environment, normalized to a range $[-1.5, 1.5]$, from the plane $z\!=\!1$ to the sphere via a stereographic projection. The target area, the T-shaped object, and the agent then lie and evolve on the sphere. As in the Euclidean case, each observation $\vo$ is composed of the $96\times 96$ RGB image of the current scene and the agent's state information on the manifold. 
All models (i.e., RFMP, SRFMP, DP) are trained for $300$ epochs in both settings. During testing, we choose the best validation epoch and roll out $500$ steps in the environment with an early stop rule terminating the execution when the coverage area is over $95\%$ of the green target area. The score for both Euclidean and Sphere Push-T tasks is the maximum coverage ratio during execution. The tests are performed with $50$ different initial states not present in the training set.

\subsubsection{Euclidean Push-T}
First, we evaluate the performance of RFMP and SRFMP in the Euclidean case for different number of function evaluations in the testing phase. The models are trained with the default parameters described in Section~\ref{subsubsec:sRFMPimplementation}.
Table~\ref{table:ODEstep euclidean pusht} shows the success rate of RFMP and SRFMP for different NFEs. We observe that both RFMP and SRFMP achieve similar success rates overall. While SRFMP demonstrates superior performance with a single NFE, RFMP achieves higher success rates with more NFEs. We hypothesize that this behavior arises from the fact that, due to the equality $\lambda_{\bm{x}}=\lambda_\tau$, the SRFMP conditional probability path resembles the optimal transport map between the prior and target distributions as in~\cite{lipmanCFM}. This, along with the stability framework of SRFMP, allows us to automatically choose the time step during inference via~\eqref{eq:SFMrecursionNsteps}, which enhances convergence in a single step.
When comparing our approaches with DP, we observe that DP performs drastically worse than both RFMP and SRFMP for a single NFE, achieving a score of only $10.9\%$. Nevertheless, the performance of DP improves when increasing the NFE and matches that of our approaches for $10$ NFE.\looseness=-1 

\begin{table}[tbp]
    \centering
    \caption{Euclidean Push-T: Impact of NFE on policies.}
    \label{table:ODEstep euclidean pusht}
    \begin{tabular}{ccccc}
       \rowcolor{gray!15}  &  \multicolumn{4}{c}{\textbf{NFE}} \\
       \rowcolor{gray!15} \multirow{-2}{*}{\textbf{Policy}}         &   $1$ &  $3$ & $5$ & $10$
    \\
       RFMP  & $0.848$ &	$\bm{0.855}$ & $\bm{0.923}$	& $\bm{0.891}$\\
     \rowcolor{gray!10}  SRFMP & $\bm{0.875}$& $0.851$ & $0.837$ & $0.856$\\
        DP & $0.109$ & $0.79$ & $0.838$ & $0.862$\\
    \end{tabular}
\end{table}

\begin{table}[tbp]
\centering
\caption{RFMP Hyperparameters Ablation on Euclidean Push-T.}
\label{table:ablationRFMPonPushT}
\resizebox{\linewidth}{!}{
\begin{tabular}{c c c c c c c}
      % \toprule
     \rowcolor{gray!15} \textbf{Parameter} & \textbf{Values} & \multicolumn{5}{c}{\textbf{Success rate}} \\ 
   \rowcolor{gray!15}   \textbf{NFE}  & & $1$ & $3$ & $5$& $10$ & $100 $ \\
      % \midrule
  \multirow{3}{*}{$T_o$}
       & $\textcolor{darkgreen}{\bm{2}}$ &$\bm{0.848}$ &$\bm{0.855}$ & $\bm{0.923}$& $\bm{0.891}$&  $\bm{0.91}$ \\
        & $ 8 $ & $ 0.195$&	$0.16$	&$0.154$	&$0.168$&	$0.179$\\
       &  $16$  &  $0.135$&	$0.143$	&$0.14$&	$0.133$&	$0.135$
\\
       % \hline
   \rowcolor{gray!5}   &  $8$  & $0.754$	&$0.835$	&$0.827$	&$0.839$	&$0.85$\\
    \rowcolor{gray!5}   &  $\textcolor{darkgreen}{\bm{16}}$  &$\bm{0.848}$ & $0.855$ & $\bm{0.923}$& $0.891$&  $0.91$ \\
     \rowcolor{gray!5}  \multirow{-3}{*}{$T_p$} &  $32$  &  $0.799$&	$\bm{0.906}$&	$0.878$	&$\bm{0.929}$	&$\bm{0.93}$\\
       % \hline
       \multirow{3}{*}{$\eta$} & $\textcolor{darkgreen}{\bm{1 \times 10^{-4}}}$ &$\bm{0.848}$ & $0.855$ & $0.923$& $0.891$&  $0.91$ \\
      &  $\num{5e-5}$  &  $0.797$&	$0.863$	&$0.843	$&$0.897$	&$0.889$\\
       &  $\num{1e-5}$  &  $0.641	$&$0.771$	&$0.805	$&$0.88$	&$0.841$ \\
       % \hline
     \rowcolor{gray!5}   & $\textcolor{darkgreen}{\bm{0.001}}$  &  $0.848$ & $0.855$ & $\bm{0.923}$& $0.891$&  $\bm{0.91}$ \\
   \rowcolor{gray!5}     &  $0.005 $ &  $0.846$&	$\bm{0.882}$	&$0.875$	&$0.866$	&$0.856$\\
     \rowcolor{gray!5}  \multirow{-3}{*}{$w_d$}  &  $0.01 $ &  $\bm{0.868}$&	$0.831$	&$0.842$	&$\bm{0.927}$&	$0.853$\\
       % \bottomrule
    \end{tabular}
    }
\end{table}

Next, we ablate the action prediction horizon $T_p$, observation horizon $T_o$, learning rate $\eta$, and weight decay $w_d$ for RFMP and SRFMP. We consider $3$ different values for each, while setting the other hyperparameters to their default values, and test the resulting models with $5$ different NFE. For SRFMP, we additionally ablate the parameters $\lambda_\vx$ and $\lambda_\tau$ for $3$ different ratios $\lambda_\vx / \lambda_\tau$ and $4$ values for each ratio. Each setup is tested with $50$ seeds, resulting in a total of $2250$ and $4000$ experiments for RFMP and SRFMP. The results are reported in Tables~\ref{table:ablationRFMPonPushT} and~\ref{table:ablationSRFMPonPushT}, respectively.
We observe that a short observation horizon $T_o=2$ leads to the best performance for both models. This is consistent with the task, as the current and previous images accurately provide the required information for the next pushing action, while the actions associated with past images rapidly become outdated. Moreover, we observe that an action prediction horizon $T_p\!=\!16$ leads to the highest score. We hypothesize that this horizon allows the model to maintain temporal consistency, while providing frequent enough updates of the actions according to the current observations. Concerning SRFMP, we find that $\lambda_\vx \!=\! \lambda_\tau \!=\! 2.5$ leads to the highest success rates. Interestingly, this choice leads to the ratio $\lambda_\vx / \lambda_\tau \!=\! 1$, in which case the flow of $\vx$ follows the Gaussian CFM~\eqref{Eq:LipmanCFM} of~\cite{lipmanCFM} with $\sigma\to0$ for $\tau\!=\![0,1]$, see~\cite[Cor 4.12]{stableflow}.
In the next experiments, we use the default parameters resulting from our ablations, i.e., $T_p\!=\!16$, $T_o\!=\!2$, $\eta\!=\!1\!\times \!10^{-4}$, $w_d\!=\!0.001$, and $\lambda_\vx \!=\! \lambda_\tau\!=\! 2.5$.

\begin{table}[tbp]
\centering
  \caption{SRFMP Hyperparameters Ablation on Euclidean Push-T.}
  \label{table:ablationSRFMPonPushT}
\begin{tabular}{c c c c c c c}
        % \toprule
      \rowcolor{gray!15}  \textbf{Parameter} & \textbf{Values} & \multicolumn{4}{c}{\textbf{Success rate}} \\ 
     \rowcolor{gray!15}   \textbf{NFE}  & & $1$ & $3$ &$ 5$& $10$  \\
        % \midrule
        \multirow{3}{*}{$T_o$} & $\textcolor{darkgreen}{\bm{2}}$  & $\bm{0.875}$&	$\bm{0.851}$	&$\bm{0.837}$&	$\bm{0.856}$ \\
         &  $8$  &  $0.124$&	$0.139$&	$0.147$&	$0.13$\\
         &  $16$  & $ 0.145$	&$0.138$&	$0.149$&	$0.144$\\
         % \hline
       \rowcolor{gray!5}   &  $8$  &  $0.816$&	$0.726$&	$0.592$&	$0.318$ \\
     \rowcolor{gray!5}     &  $\textcolor{darkgreen}{\bm{16}}$  & $\bm{0.875}$&	$0.851$	&$0.837$&	$\bm{0.856}$ \\
 \rowcolor{gray!5}       \multirow{-3}{*}{$T_p$}  &  $32$  &  $0.852$&	$\bm{0.861}$&	$\bm{0.881}$	&$0.829$ \\
         % \hline
         \multirow{3}{*}{$\eta$} & $\textcolor{darkgreen}{\bm{1.0 \times 10^{-4}}}$  &$\bm{0.875}$&	$\bm{0.851}$	&$\bm{0.837}$&$\bm{0.856}$ \\
          &  $\num{5e-5}$  & $ 0.754$	&$0.621$	&$0.456$&	$0.334$ \\
           &  $\num{1e-5}$  & $ 0.602$	&$0.56$	&$0.443$&	$0.288$ \\
           % \hline
      \rowcolor{gray!5}    & $\textcolor{darkgreen}{\bm{0.001}}$  & $\bm{0.875}$ &	$\bm{0.851}$	&$\bm{0.837}$&$\bm{0.856}$ \\
    \rowcolor{gray!5}     &  $0.005$  & $0.837$&	$0.826$&	$0.826$&	$0.684$  \\
   \rowcolor{gray!5}  \multirow{-3}{*}{$w_d$}    &  $0.01$  &  $0.733$&	$0.75$	&$0.768	$&$0.571$ \\
         % \hline
         \multirow{12}{*}{$\lambda_{\vx} \quad \lambda_\tau$} & $1$ $0.2$ & $0.74$&$0.614$&	$0.494$&	$0.457$ \\
        & $1$ $1$ & $0.85$& 	$0.777$	& $0.608$	& $0.416$\\
        & $1$ $6$ & $0.87$&	$0.768$&	$0.753$&	$0.812$ \\
        & $2.5$ $0.2$ & $0.832$ & $0.392$ & $0.576$ & $0.549$ \\ 
       &  $\textcolor{darkgreen}{\bm{2.5}}$ $\textcolor{darkgreen}{\bm{2.5}}$ & $\bm{0.875}$&	$\bm{0.851}$	&$0.837$&	$\bm{0.856}$\\
         & $2.5$ $15$ & $0.832$ & $0.799$ & $0.789$ & $0.807$ \\
         & $5$ $1$ & $0.796$& 	$0.741$& 	$0.614$	& $0.513$\\
         & $5$ $5$ & $0.845$&	$0.825$	&$0.797$	&$0.642$ \\
         & $5$ $30$  & $0.822$	 &$0.830$ &	$0.772$ &	$0.743$\\
         & $7.5$ $1.5$ & $0.799$	&$0.685$	&$0.442$	&$0.456$\\
         & $7.5$ $7.5$ &  $0.787$	&$0.8$	&$0.809$	&$0.633$\\
         & $7.5$ $45$ & $0.782$&	$0.817$&	$\bm{0.844}$	& $0.814$\\
        % \bottomrule
    \end{tabular}
\end{table} 

\subsubsection{Sphere Push-T}
Next, we test the ability of RFMP and SRFMP to generate motions on non-Euclidean manifolds with the \textsc{Sphere Push-T} task.
We evaluate two types of Riemannian prior distributions for RFMP and SRFMP, namely a spherical uniform distribution and wrapped Gaussian distribution (see Figure~\ref{figure:sphere_prior}). We additionally consider a Euclidean Gaussian distribution for DP. Notice that the actions generated by DP are normalized in a post-processing step to ensure that they belong to the sphere.
The corresponding performance are reported in Table~\ref{table:sphere pusht}. Our results indicate that the choice of prior distribution significantly impacts the performance of both RFMP and SRFMP. Specifically, we observe that RFMP and SRFMP with a uniform sphere distribution consistently outperform their counterparts with wrapped Gaussian distribution. We hypothesize that RFMP or SRFMP benefit from having samples that are close to the data support, which leads to simpler vector fields to learn. In other words, uniform distribution provides more samples around the data distribution, which potentially lead to simpler vector fields. DP exhibit poor performance with sphere-based prior distributions, suggesting its ineffectiveness in handling such priors. Instead, DP's performance drastically improves when using a Euclidean Gaussian distribution and higher NFE. Note that this high performance does not scale to higher dimensional settings as already evident in the real-world experiments reported in Section~\ref{subsec:realExp}, where the effect of ignoring the geometry of the parameters exacerbates, which is a known issue when naively operating with Riemannian data~\cite{Jaquier24:Fallacy}. Importantly, SRFMP is consistently more robust to NFE and achieves high performance with a single NFE, leading to shorter inference times for similar performance compared to RFMP and DP. 

\begin{table}[tbp]
    \caption{Impact of the prior distribution on Sphere Push-T Task.}
    \label{table:sphere pusht}
    \centering
    \begin{tabular}{ccccc}
    % \toprule
   \rowcolor{gray!15}   & \multicolumn{4}{c}{\textbf{NFE}} \\
     \rowcolor{gray!15}    \multirow{-2}{*}{\textbf{Policy}}                    & $1$ & $3$ & $5$ & $10$
    \\
     % \midrule
       RFMP sphere uniform  &$\bm{0.871}$&	$0.746$&	$0.77$&	$0.817$\\
       RFMP sphere Gaussian &$0.587$&	$0.724$&	$0.748$&	$0.733$ \\
       \rowcolor{gray!5} SRFMP sphere uniform &$0.772$	&$0.736$&	$0.796$&	$0.829$\\
      \rowcolor{gray!5}  SRFMP sphere Gaussian & $0.707$	&$0.706$	&$0.735$&	$0.707$ \\
        DP sphere uniform & $0.274$ &	$0.261$ &	$0.235$ &	$0.197$\\
        DP sphere Gaussian & $0.170$	&$0.162$&	$0.231$&	$0.227$ \\
        DP euclidean Gaussian & $0.227$ & $\bm{0.796}$ & $\bm{0.813}$ & $\bm{0.885}$ \\
         % \bottomrule
    \end{tabular}
\end{table}

\begin{table}[tbp]
    \centering
    \caption{Influence of Integration Time on RFMP and SRFMP.}
    \label{table:time boudary influence}
    \begin{tabular}{cccc}
     \rowcolor{gray!10} \textbf{Euclidean \textsc{Push-T}} & $t=1.0$ & $t=1.2$ & $t=1.6$ \\
     RFMP & $0.855$ & $0.492$ & $0.191$ \\
     SRFMP & $\bm{0.862}$ & $\bm{0.851}$ & $\bm{0.829}$ \\
     \rowcolor{gray!10} \textbf{Sphere \textsc{Push-T}} &$t=1.0$ & $t=1.2$ & $t=1.6$ \\
     RFMP & $\bm{0.736}$ & $0.574$ & $0.264$ \\
     SRFMP & $0.727$ & $\bm{0.736}$ & $\bm{0.685}$ 
    % \bottomrule
    \end{tabular}
\end{table}
\begin{figure}[t]
\centering
\includegraphics[width=0.32\linewidth]{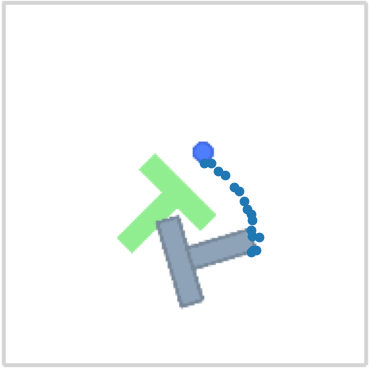} 
\includegraphics[width=0.32\linewidth]{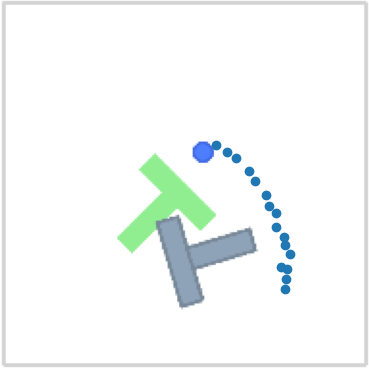} 
\includegraphics[width=0.32\linewidth]{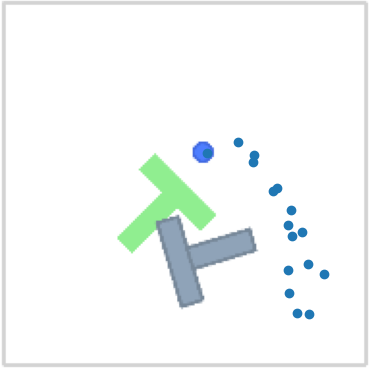} \\
\includegraphics[width=0.32\linewidth]{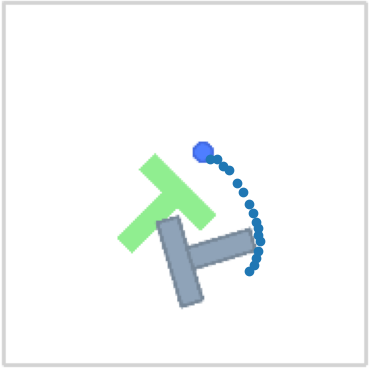} 
\includegraphics[width=0.32\linewidth]{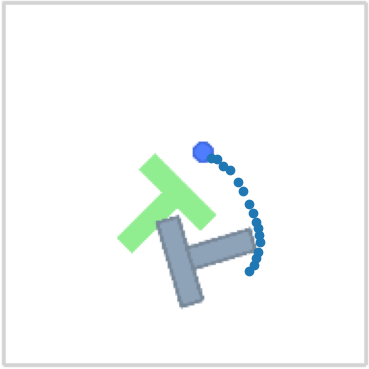} 
\includegraphics[width=0.32\linewidth]{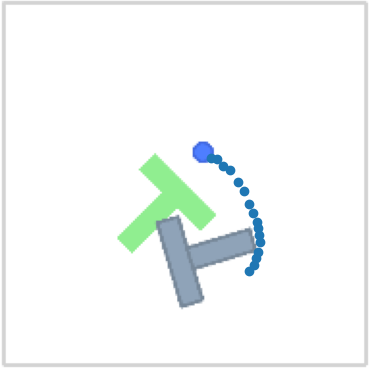} \\
\vspace{1em}
\includegraphics[width=0.32\linewidth]{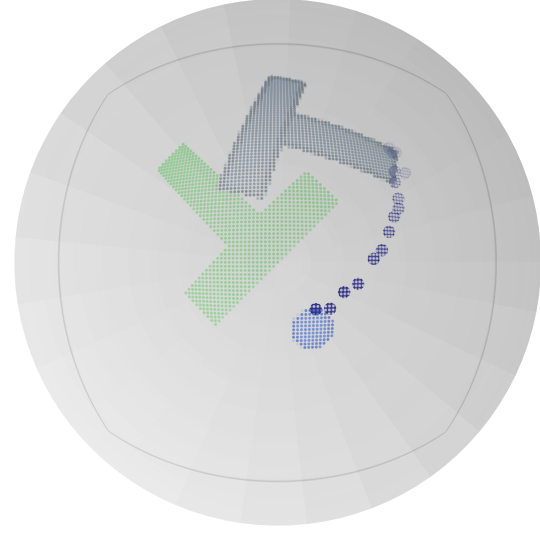} 
\includegraphics[width=0.32\linewidth]{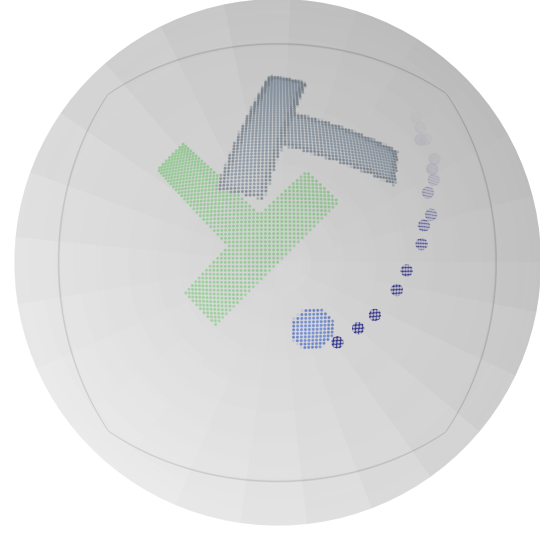} 
\includegraphics[width=0.32\linewidth]{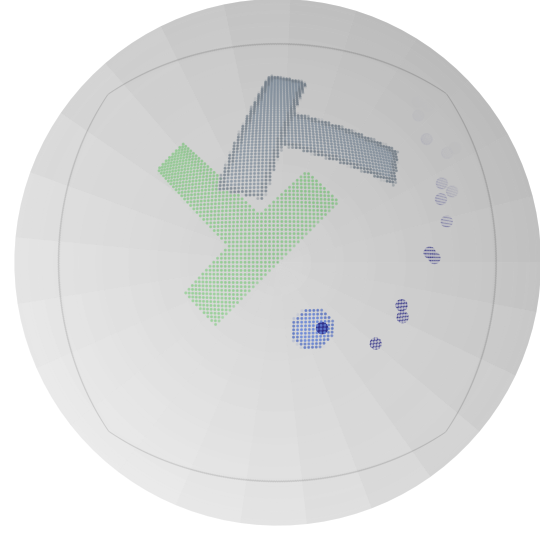} \\
\includegraphics[width=0.32\linewidth]{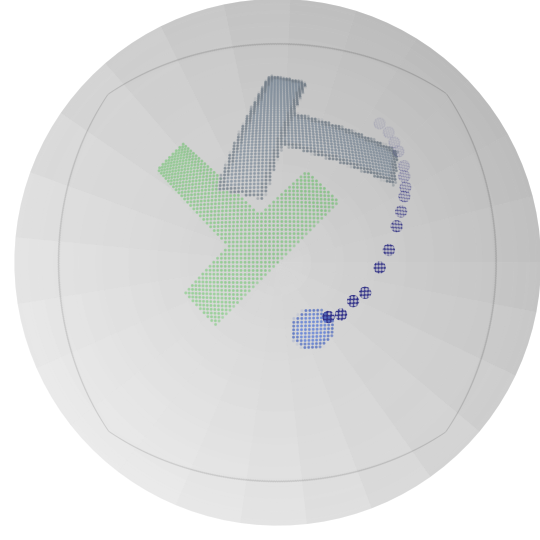} 
\includegraphics[width=0.32\linewidth]{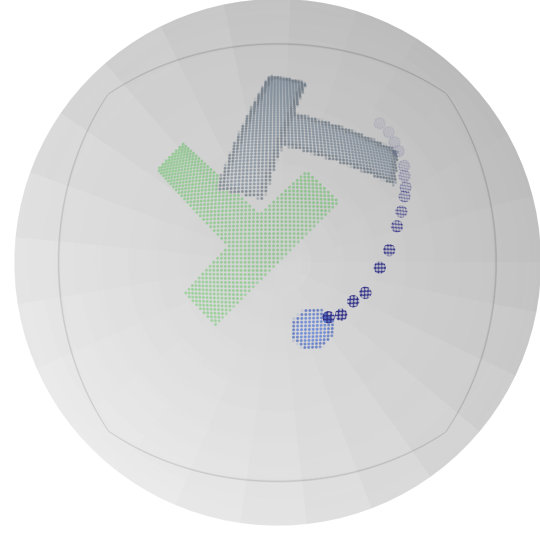} 
\includegraphics[width=0.32\linewidth]{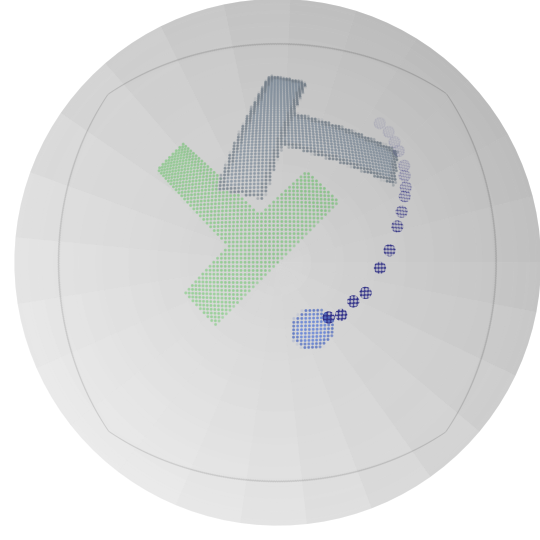} \\
\caption{Action series generated by RFMP (\textbf{first and third rows}) and SRFMP (\textbf{second and fourth rows}) trained on the \textsc{Push-T} tasks at integration times $t=\{0.8, 1.2, 1.6\}$. }
\label{fig:influence of time bound}
\end{figure}

\subsubsection{Influence of Integration Time Boundary}
We further assess the robustness of SRFMP to varying time boundaries on the \textsc{Push-T} tasks by increasing the time boundary during inference. The performance of both RFMP and SRFMP is summarized in Table~\ref{table:time boudary influence} with result presented for $\text{NFE}\!=\!3$ under the time boundaries $t\!=\!1$ and $t\!=\!1.2$, as well as for $\text{NFE} =4$ under the time boundaries $t\!=\!1.6$. Our results show that the performance of RFMP is highly sensitive to the time boundary, gradually declining as the boundary increases. In contrast, SRFMP demonstrates remarkable robustness, with minimal variation across different time boundaries. As illustrated in Figure~\ref{fig:influence of time bound}, the quality of action series generated by RFMP noticeably deteriorates with increasing time boundaries, whereas SRFMP consistently delivers high-quality action series regardless of the time boundary. 

\subsection{Robomimic Benchmark}
\label{subsec:simulatedExp}
Next, we evaluate RFMP and SRFMP on the well-known Robomimic robotic manipulation benchmark~\cite{robomimic}. This benchmark consists of five tasks with varying difficulty levels for which it provides two types of demonstrations, namely proficient human (PH) high-quality teleoperated demonstrations, and mixed human (MH) demonstrations. Each demonstration contains multi-modal observations, including state information, images, and depth data. We report results on five tasks (\textsc{Lift}, \textsc{Can}, \textsc{Square}, \textsc{Tool Hang}, and \textsc{Transport}) from the Robomimic dataset with $200$ PH demonstrations for training for both state- and vision-based observations. Note that the difficulty of the selected tasks becomes progressively more challenging. The score of each of the $50$ trials is determined by whether the task is completed successfully after a given number of steps ($300$ for \textsc{Lift}, $500$ for \textsc{Can} and \textsc{Square}, $700$ for \textsc{Tool Hang}, and $500$ for \textsc{Transport}). The performance is then the percentage of successful trials.

\begin{figure}[tbp]
\centering
\includegraphics[width=\linewidth]{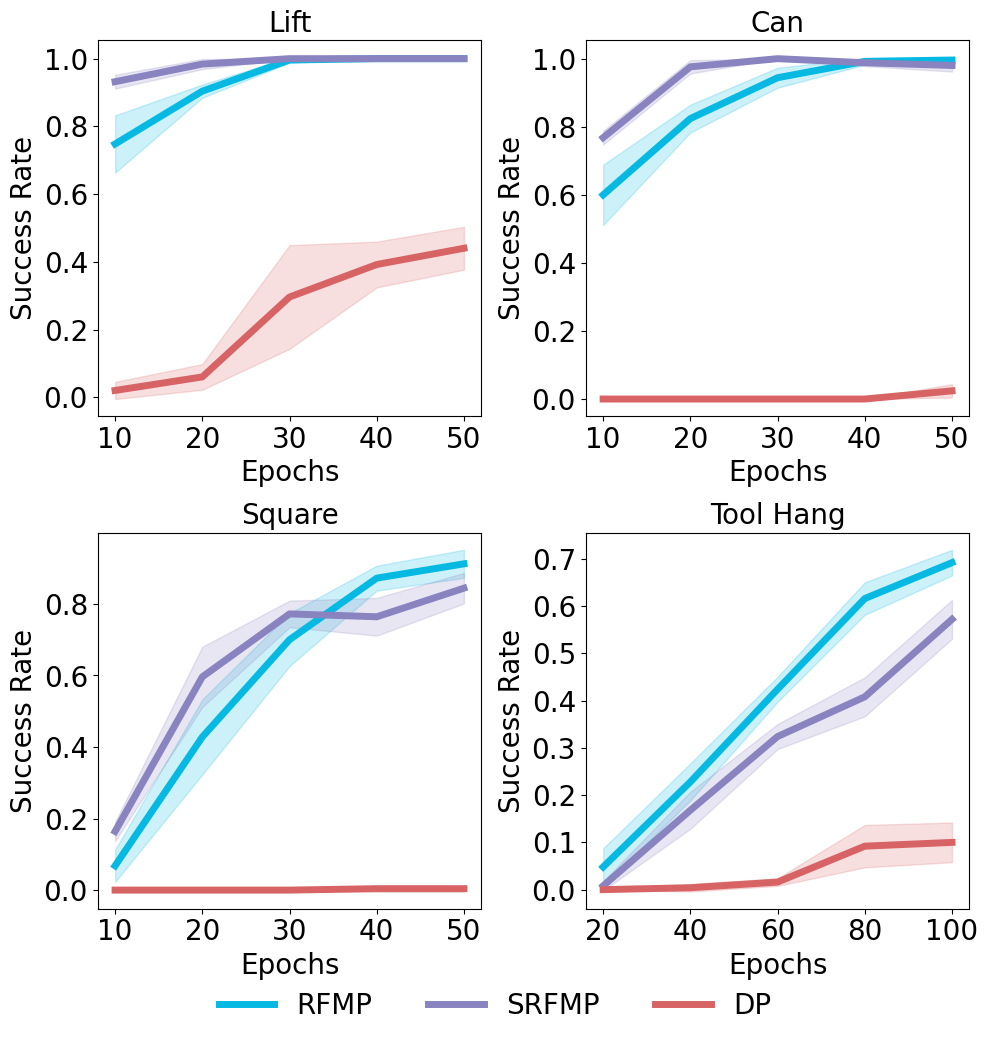} 
\caption{Success rate (mean and standard deviation) on Robomimic tasks with state-based observations at different checkpoints. The models performance of \textsc{Lift}, \textsc{Can}, and \textsc{Square} tasks is checked every $10$ epochs throughout the $50$-epoch training process using $3$ NFE. For the \textsc{Tool Hang} task, the models are trained over $100$ epochs and checked every $20$ epochs using $10$ NFE.}
\label{fig:robomimic epoch influence}
\end{figure}

\subsubsection{State-based Observations}
We first assess the training efficiency of RFMP and SRFMP and compare it against DP by analyzing their performance at different training stages. Figure~\ref{fig:robomimic epoch influence} shows the success rate of the three policies as a function of the number of training epochs for the tasks \textsc{Lift}, \textsc{Can}, \textsc{Square}, and \textsc{Tool Hang}.
All policies are evaluated with $3$ NFE for \textsc{Lift}, \textsc{Can}, and \textsc{Square}, and with $10$ NFE for \textsc{Tool Hang}.
We observe that RFMP and SRFMP consistently outperform DP across all tasks, requiring fewer training epochs to achieve comparable or superior performance. For the easier tasks (\textsc{Lift} and \textsc{Can}), both RFMP and SRFMP achieve high performance after just $20$ training epochs, while the success rate of DP remains low after $50$ epochs. This trend persists in the harder tasks (\textsc{Square} and \textsc{Tool Hang}), with RFMP and SRFMP reaching high success rates significantly faster than DP. \looseness-1

\begin{table*}[tbp]
    \fontsize{6pt}{6.5pt}\selectfont
    \centering
    \belowrulesep=0pt
    \aboverulesep=0pt
    \caption{Success rate as a function of different NFE values on the state-based Robomimic tasks.}
    \resizebox{.75\textwidth}{!}{
    \begin{tabular}{c|ccccc}
    % \toprule
     \rowcolor{gray!15}   & \multicolumn{5}{c}{\textsc{\textbf{Lift}}} 
    \\
   \rowcolor{gray!15}  \textbf{NFE}   & $1$ & $2$ & $3$ & $5$ & $10$ \\
     % \midrule
        RFMP  & $0.992 \pm 0.010$&	$\bm{0.992 \pm 0.010}$&	$\bm{0.992 \pm 0.010}$	&$0.992 \pm 0.010$&	$0.992 \pm 0.010$ 
\\
      \rowcolor{gray!5} SRFMP &$\bm{1.000 \pm 0.000}$	&$\bm{0.992 \pm 0.010}$&	$\bm{0.992 \pm 0.010}$&	$\bm{0.996 \pm 0.008}$	&$\bm{1.000 \pm 0.000}$
       \\
       DP & $0.008 \pm 0.010$&	$0.756 \pm 0.034$&	$0.948 \pm 0.016$&	$0.956 \pm 0.015$	&$0.976 \pm 0.015$
        \\
    \rowcolor{gray!15}   & \multicolumn{5}{c}{\textsc{\textbf{Can}}} 
    \\
   \rowcolor{gray!15}  \textbf{NFE}    & $1$ & $2$ & $3$ & $5$ & $10$\\
     % \midrule
        RFMP  & 
        $0.976 \pm 0.023$&	$\bm{0.996 \pm 0.008}$&	$\bm{0.996 \pm 0.008}$	&$\bm{0.996 \pm 0.008}$	&$\bm{0.996 \pm 0.008}$
\\
      \rowcolor{gray!5} SRFMP &
       $\bm{0.980 \pm 0.022}$&	$0.980 \pm 0.013$&	$\bm{0.996 \pm 0.008}$	&$0.992 \pm 0.010$&	$0.968 \pm 0.020$
       \\
       DP &  
        $0.004 \pm 0.008$	&$0.340 \pm 0.013$&	$0.836 \pm 0.020$&	$0.924 \pm 0.015$	&$0.908 \pm 0.010$
        \\
        \rowcolor{gray!15} & \multicolumn{5}{c}{\textsc{\textbf{Square}}} \\
    \rowcolor{gray!15}   \textbf{NFE} & $1$ & $2$ & $3$ & $5$ & $10$  \\
    RFMP & $\bm{0.792 \pm 0.027}$ & $\bm{0.848 \pm 0.020}$ & $\bm{0.920 \pm 0.018}$ & $\bm{0.896 \pm 0.041}$ & $\bm{0.912 \pm 0.016}$  \\
    \rowcolor{gray!5} SRFMP & $0.776 \pm 0.029$ & $0.800 \pm 0.052$ & $0.828 \pm 0.016$ & $0.824 \pm 0.008$ & $0.848 \pm 0.030$ \\
    DP & $0.012 \pm 0.010$ & $0.384 \pm 0.023$ & $0.628 \pm 0.016$ & $0.672 \pm 0.020$ & $0.684 \pm 0.015$  \\
    \rowcolor{gray!15}    & \multicolumn{5}{c}{\textsc{\textbf{Tool Hang}}} \\
    \rowcolor{gray!15}   \textbf{NFE} & $1$ & $2$ & $3$ & $5$ & $10$ \\
    RFMP & 
          $0.152 \pm 0.020$ & $\bm{0.316 \pm 0.023}$ & $\bm{0.368 \pm 0.032}$ & $\bm{0.572 \pm 0.027}$ & $\bm{0.716 \pm 0.023}$ \\
    \rowcolor{gray!5} SRFMP & 
         $\bm{0.240 \pm 0.028}$ & $0.224 \pm 0.020$ & $0.308 \pm 0.020$ & $0.516 \pm 0.023$ & $0.568 \pm 0.035$ \\
    DP & 
        $0.000 \pm 0.000$ & $0.008 \pm 0.010$ & $0.008 \pm 0.010$ & $0.092 \pm 0.024$ & $0.092 \pm 0.016$ \\
         % \bottomrule
    \end{tabular}}
    \label{table:robomimic}
\end{table*}
\begin{table*}[tbp]
    \fontsize{7pt}{8pt}\selectfont
    \centering
    \belowrulesep=0pt
    \aboverulesep=0pt
    \caption{Jerkiness of predicted robot trajectories for different NFE on the robomimic tasks with state-based observations. All values are expressed in thousands, where the lower the smoother the prediction.}
    \resizebox{.63\textwidth}{!}{
    \begin{tabular}{c|ccccc}
    % \toprule
    \rowcolor{gray!15}  &  \multicolumn{5}{c}{\textsc{\textbf{Lift}}}
    \\
  \rowcolor{gray!15} \textbf{NFE}      & $1$ & $2$ & $3$ & $5$ & $10$ \\
     % \midrule
       RFMP  & $\bm{9.97 \pm 1.23}$ &	$\bm{8.80 \pm 0.19}$ &	$9.06 \pm 0.08$ &	$8.96 \pm 0.08$ &	$8.49 \pm 0.21$  
\\
    \rowcolor{gray!5}     SRFMP & $10.32 \pm 0.41$ &	$9.90 \pm 031$&	$\bm{8.00 \pm 0.14}$&	$9.39 \pm 0.23$	&$9.03 \pm 0.19$
       \\
        DP & $329.20 \pm 47.44$&	$13.96 \pm 0.89$&	$8.06 \pm 0.42$&	$\bm{7.48 \pm 0.13}$&	$\bm{5.9 \pm 0.23}$
        \\
    \rowcolor{gray!15} &  \multicolumn{5}{c}{\textsc{\textbf{Can}}} 
    \\
  \rowcolor{gray!15} \textbf{NFE}     & $1$ & $2$ & $3$ & $5$ & $10$\\
     % \midrule
       RFMP  &
       $7.88 \pm 0.14$ &	$\bm{6.38 \pm 0.40}$ &	$\bm{6.37 \pm 0.17}$ &	$6.40 \pm 0.07$ &	$7.10 \pm 1.30$ 
       
\\
    \rowcolor{gray!5}     SRFMP &
$\bm{7.64 \pm 0.23}$&	$7.63 \pm 0.18$&	$6.69 \pm 0.27$&	$6.98 \pm 0.13$&	$6.94 \pm 0.08$

       \\
        DP &
$526.00 \pm 65.50$&	$18.98 \pm 2.82$	&$6.68 \pm 0.25$&	$\bm{6.28 \pm 0.35}$	&$\bm{6.22 \pm 0.16}$
        \\
        \rowcolor{gray!15}   & \multicolumn{5}{c}{\textsc{\textbf{Square}}} \\
    \rowcolor{gray!15}   \textbf{NFE} & $1$ & $2$ & $3$ & $5$ & $10$ \\
    RFMP & $\bm{9.14 \pm 0.19}$ &	$\bm{6.22 \pm 0.29}$ &	$\bm{5.54 \pm 0.24}$&	$\bm{5.17 \pm 0.16}$ &	$\bm{4.36 \pm 0.30}$ \\
       \rowcolor{gray!5} SRFMP & $9.16 \pm 0.31$	&$9.43 \pm 0.16$&	$7.28 \pm 0.97$&	$7.70 \pm 0.50$&	$7.74 \pm 0.26$\\
    DP & $548.20 \pm 54.77$&	$20.26 \pm 2.47$	&$7.24 \pm 0.54$&	$10.02 \pm 2.64$	&$7.62 \pm 1.44$\\
    \rowcolor{gray!15}   &  \multicolumn{5}{c}{\textsc{\textbf{Tool Hang}}} \\
        \rowcolor{gray!15}   \textbf{NFE} & $1$ & $2$ & $3$ & $5$ & $10$ \\
        RFMP & 
           $\bm{4.49 \pm 0.35}$&	$\bm{4.26 \pm 0.19}$&	$\bm{4.65 \pm 0.32}$&	$\bm{5.07 \pm 0.19}$&	$\bm{4.66 \pm 0.31}$ \\
           \rowcolor{gray!5} SRFMP & 
    $5.24 \pm 0.18$&	$5.10 \pm 0.20$&	$5.23 \pm 0.38$&	$5.62 \pm 0.42$&	$5.62 \pm 0.40$ \\
        DP & 
    $699.40 \pm 90.91$&	$9.52 \pm 0.30$& $8.79 \pm 1.51$& $7.12 \pm 1.94$&	$6.96 \pm 1.35$
         % \bottomrule
    \end{tabular}}
    \label{table:robomimic smoothness}
\end{table*}

Next, we evaluate the performance of the policies for different NFE in the testing phase. For RFMP and SRFMP, we use the $50$-epoch models for \textsc{Lift}, \textsc{Can}, and \textsc{Square}, and the $100$-epoch models for \textsc{Tool Hang}. DP is further trained for a total of $300$ epochs and we select the model at the best validation epoch. The results are reported in Table~\ref{table:robomimic}. RFMP and SRFMP outperform DP for all tasks and all NFE, even though DP was trained for more epochs. Moreover, we observe that RFMP and SRFMP are generally more robust to low NFE than DP. They achieve $100$\% success rate at almost all NFE values for the easier \textsc{Lift} and \textsc{Can} tasks, while DP's performance drastically drops for $1$ and $2$ NFE. We observe a similar trend for the \textsc{Square} task, where the performance of RFMP and SRFMP slightly improves when increasing the NFE. The performance of all models drops for \textsc{Tool Hang}, which is the most complex of the four considered tasks. In this case, the performance of RFMP and SRFMP is limited for low NFE values and improves for higher NFE. DP performs poorly for all considered NFE values. Table~\ref{table:robomimic smoothness} reports the jerkiness as a measure of the smoothness of the trajectories generated by the different policies. We observe that RFMP and SRFMP produce arguably smoother trajectories than DP for low NFE, as indicated by the lower jerkiness values. The smoothness of the trajectories becomes comparable for higher NFE. In summary, both RFMP and SRFMP achieve high success rates and smooth action predictions with low NFE, enabling faster inference without compromising task completion.
\begin{table}[tbp]
\centering
\caption{Training time (in seconds) per epoch on Robomimic vision-based \textsc{Lift} and \textsc{Square}.}
\label{table:train time}
\resizebox{\linewidth}{!}{
\begin{tabular}{ccccc}
% \toprule
\rowcolor{gray!15} \textbf{Task} & \textbf{RFMP} & \textbf{SRFMP} & \textbf{DP} & \textbf{CP}
\\
% \midrule
\textsc{Lift} &
$17.49 \pm 0.03$ &
$17.63 \pm 0.15$ &
$16.07 \pm 0.20$ &
$31.44 \pm 0.27$ \\
\rowcolor{gray!5} \textsc{Square} &
$57.62 \pm 0.30$ &
$56.67 \pm 0.18$ &
$53.16 \pm 0.16$ &
$102.86 \pm 0.61$ \\

% \bottomrule
\end{tabular}}
\end{table}

\subsubsection{Vision-based Observations}
Next, we assess our models performance when the vector field is conditioned on visual observations. We consider the tasks \textsc{Lift}, \textsc{Can}, \textsc{Square}, and \textsc{Transport} with the same policy settings and networks (see Table~\ref{table:hyperParamRFMP})\footnote{Note that we omit \textsc{Tool Hang} due to its significant computational cost.}. Each observation $\vo^s$ corresponds to a visual embedding derived from camera images. For \textsc{Lift}, \textsc{Can}, and \textsc{Square}, we use one over-the-shoulder and one in-hand camera. For \textsc{Transport}, which involves bimanual manipulation, we use two in-hand cameras and two over-the-shoulder cameras. We train the models for a total $100$ epochs for \textsc{Lift}, \textsc{Can}, and \textsc{Square} and for $200$ epochs for \textsc{Transport} and use the best-performing checkpoint for evaluation.

The performance of different policies is reported in Table~\ref{table:robomimic vision}. RFMP and SRFMP consistently outperform DP and CP on all tasks, regardless of the NFE. As for the previous experiments, our models are remarkably robust to changes in NFE compared to DP. Importantly, SRFMP consistently outperforms RFMP for $1$ and $2$ NFE. Regarding \textsc{Can} and \textsc{Square} tasks, SRFMP with $1$ NFE achieved performance on par with RFMP using $3$ NFE. This efficiency gain showcases the benefits of enhancing the policies with stability to the target distribution for reducing their inference time.

In contrast to CP, RFMP and SRFMP rely on a simple, single-stage training pipeline, thus featuring easier and faster training in addition to fast inference. The training time per epoch of each policy on task \textsc{Lift} and \textsc{Square} tasks is summarized in Table~\ref{table:train time}. All experiments were conducted on an NVIDIA RTX 4060Ti GPU using the same batch size across policies. The result indicate that RFMP, SRFMP, and DP have comparable per-epoch training times. In contrast, CP requires approximately twice as much time per epoch, primarily due to its training procedure involving distillation from a teacher policy. When accounting for the additional time required to pretrain DP the total training time per epoch for CP becomes roughly three times that of the other methods.

\begin{table*}[tbp]
    \caption{Success rate as a function of NFE on vision-based robomimic tasks. 
    }
    \label{table:robomimic vision}
    \centering
    \setlength{\tabcolsep}{3.5pt}
    \belowrulesep=0pt
    \aboverulesep=0pt
    \begin{tabular}{c|ccccc|ccccc|ccccc|ccccc}
    % \toprule
 \rowcolor{gray!15} \textbf{Task}     &  \multicolumn{5}{c|}{\textsc{\textbf{Lift}}} 
      &  \multicolumn{5}{c|}{\textsc{\textbf{Can}}} 
      &  \multicolumn{5}{c}{\textsc{\textbf{Square}}} 
      &  \multicolumn{5}{c}{\textsc{\textbf{Transport}}}
    \\
    % \hline
   \rowcolor{gray!15}  \textbf{NFE}   & $1$ & $2$ & $3$ & $5$ & $10$ & $1$ & $2$ & $3$ & $5$ & $10$ & $1$ & $2$ & $3$ & $5$ & $10$  & $1$ & $2$ & $3$ & $5$ & $10$\\
     % \midrule
       RFMP  & $\bm{1}$&	$\bm{1}$&	$\bm{1}$	&$\bm{1}$ &	$\bm{1}$ &$0.78$&	$0.82$&	$\bm{0.9}$&	$\bm{0.96}$&	$\bm{0.94}$
        &$0.56$&	$0.74$&	$\bm{0.9}$&	$\bm{0.9}$&	$\bm{0.9}$ &
        $0.6$ & \bm{$0.82$} & $0.78$ & \bm{$0.78$} & $0.8$
\\
       \rowcolor{gray!5}  SRFMP &$\bm{1}$	&$\bm{1}$&	$\bm{1}$&	$\bm{1}$	&$\bm{1}$
       &$\bm{0.88}$ &	$\bm{0.88}$ &	$\bm{0.9}$ &	$0.9$	 &$0.86$
       &$\bm{0.86}$&	$\bm{0.82}$&	$\bm{0.9}$&	$0.88$&	$\bm{0.9}$ &
       \bm{$0.8$} & \bm{$0.82$} & \bm{$0.82$} & \bm{$0.78$} & $0.78$
       \\
        DP & $0$&	$0.7$&	$0.96$&	$0.98$&	$0.98$ 
       &$0$ &	$0.38$	 &$0.66$ &	$0.68$ &	$0.66$ 
       &$0$	 &$0.04$ &	$0.16$ &	$0.26$	 &$0.12$ &
       $0$ & $0.54$ & $0.66$ & $0.78$ & \bm{$0.82$}
        \\
       \rowcolor{gray!5} CP & $0.5$ & $0.44$ & $0.46$ & $0.46$ & $0.4$
        & $0.82$ & $0.82$ & $0.84$ & $0.84$ & $0.82$
        & $0.32$& $0.52$ & $0.54$ & $0.48$ & $0.52$ &
        $0.4$ & $0.36$ & $0.44$ & $0.38$ & $0.36$
 \\
         % \bottomrule
    \end{tabular}
\end{table*} 

\subsection{Franka Kitchen Benchmark}

\begin{table}[tbp]
\centering
\caption{Success rate as a function of NFE on the Franka kitchen benchmark.}
\label{table:kitchen_reward}
\begin{tabular}{cccccc}
% \toprule
\rowcolor{gray!15}  & \multicolumn{5}{c}{\textbf{NFE}} \\
\rowcolor{gray!15}  \multirow{-2}{*}{\textbf{Policy}} & \textbf{1} & \textbf{2} & \textbf{3} & \textbf{5} & \textbf{10} \\
% \midrule
RFMP & $0.04$ & $0.08$ & $0.12$ & $0.14$ & $0.14$ \\
\rowcolor{gray!5} SRFMP  & $\bm{1}$ & $\bm{1}$ & $\bm{1}$ & $\bm{1}$ & $\bm{1}$ \\
DP & $0$ & $0$ & $0$ & $0.02$ & $0.02$ \\

% \bottomrule
\end{tabular}
\end{table}

We evaluate the capability of RFMP and SRFMP in handling more complex, long-horizon manipulation tasks on the widely-used \textsc{Franka Kitchen} benchmark~\cite{d4rl}. We use the dataset from~\cite{minari}, which comprises $19$ expert demonstrations totaling $4209$ time steps and involves a sequential execution of four tasks: open the microwave, put the kettle on the top burner, switch on the light and slide the cabinet. The original dataset considers actions in Euclidean space as joint angular velocities. To evaluate the performance of our Riemannian policy framework, we instead consider end-effector trajectories obtained from joint configurations via forward kinematics. Therefore, each data point is composed of the end-effector position and orientation (as a unit quaternion), and the gripper state. Consequently, the predicted action sequence lies on the product of manifolds $\euclideanspace^{3} \times \mathcal{S}^3 \times \euclideanspace^{2}$. 

Each model is trained for $500$ epochs with a batch size of $32$, and evaluated using the best validation checkpoint over $50$ test episodes with randomized initial states. The performance of different policies is reported in Table~\ref{table:kitchen_reward}. SRFMP exhibits consistently high performance across all NFE, indicating strong robustness, and significantly outperforms RFMP and DP. The low success rate of RFMP and DP is due to the fact that their trajectories often exhibit local inaccuracies, leading to failing at least one subtask. This is then reported as a failure for the long-horizon task, leading to low success rates overall. In contrast, the trajectories produced by SRFMP are more precise, leading to the successful completion of all subtasks.
 
\subsection{Real Robotic Experiments}
\label{subsec:realExp}
Finally, we evaluate RFMP and SRFMP on two real-world tasks, namely \textsc{Pick \& Place} and \textsc{Mug Flipping}, with a $7$-DoF robotic manipulator.

\subsubsection{Experimental Setup}
\begin{figure}[t]
    \centering
    \includegraphics[width=0.9\linewidth]{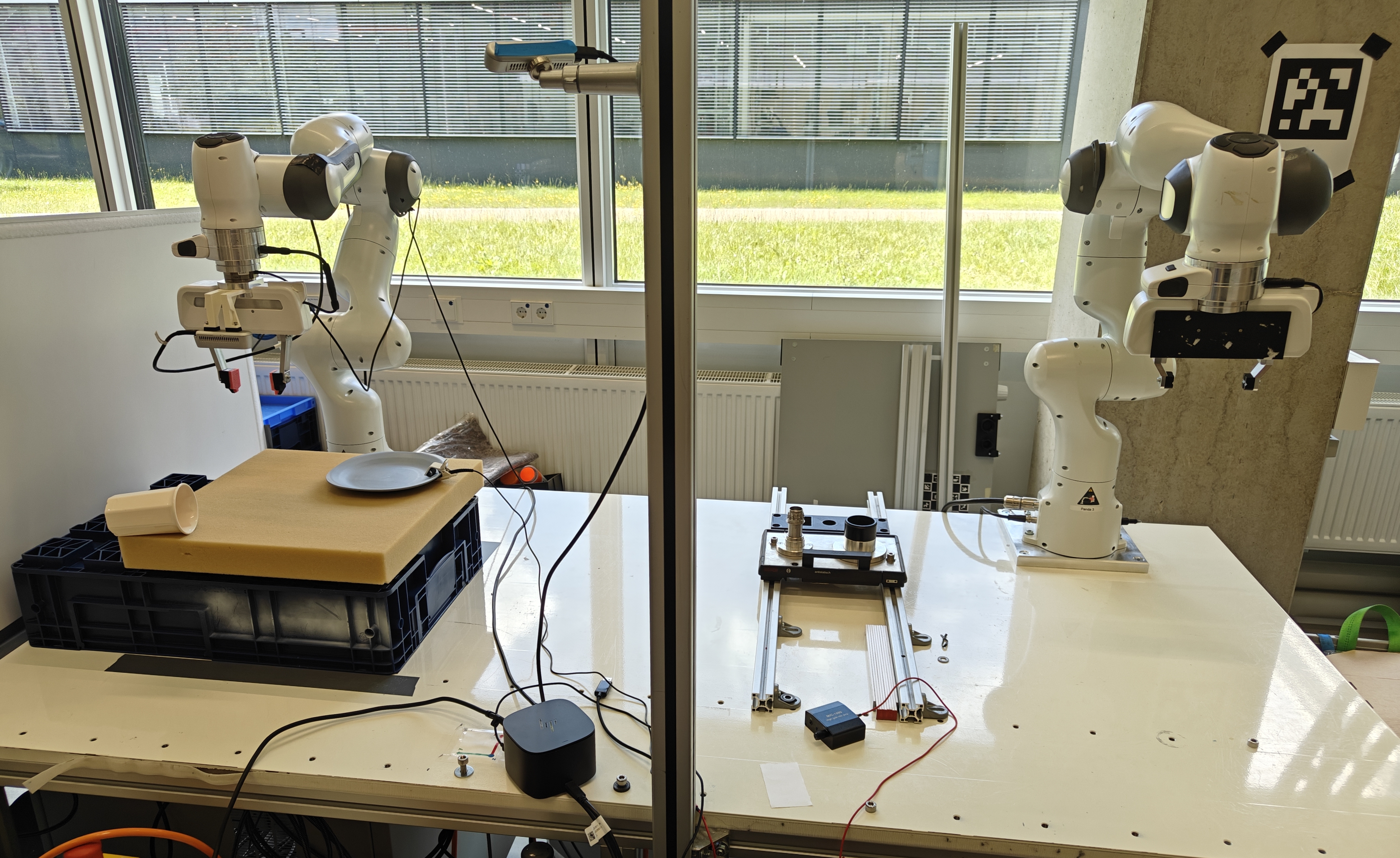}
    \caption{Robotic experimental setup consisting of two Franka Emika Panda robot arms and an over-the-shoulder camera (Realsense d435). The left arm is the \emph{target} robot, while the right one acts as the \emph{source}. During the teaching phase, a human expert controls the source arm to teleoperate the target robot. During testing, only the target arm is operational.}
    \label{fig:robotplatform}
\end{figure}

\begin{figure}[tb]
    \centering
    \includegraphics[width=0.32\linewidth]{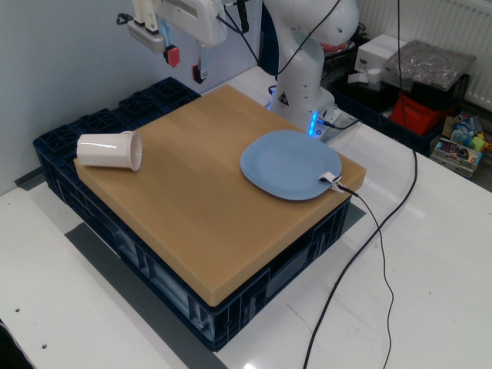}
    \hfill
    \includegraphics[width=0.32\linewidth]{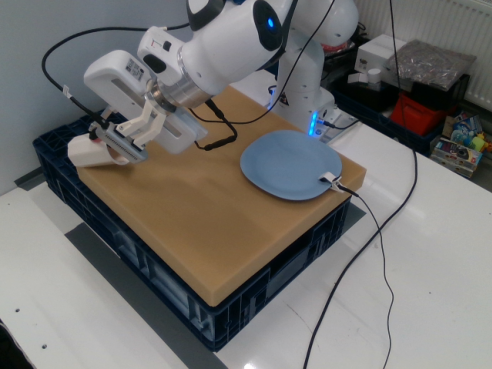}
    \hfill
    \includegraphics[width=0.32\linewidth]{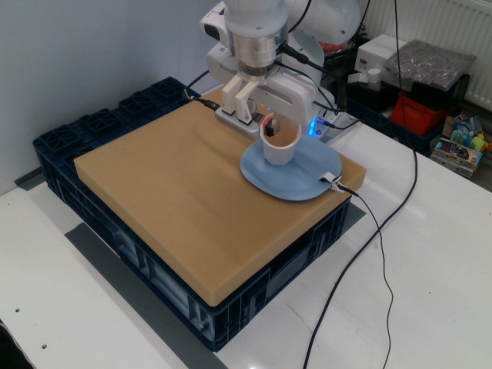}
    \caption{\textsc{Mug flipping}: The robot rotates its end-effector to align with the white mug's orientation and subsequently grasps it. It then place the mug upright on the blue plate.}
    \label{fig:rotatemug}
\end{figure}

Figures~\ref{fig:robotplatform}-\ref{fig:rotatemug} show our experimental setup. The tasks are performed on a Franka Emika Panda robot arm. We collect the demonstrations via a teleoperation system made of two robot twins. Demonstrations are collected by an expert guiding the source robot. 
The target robot reads the end-effector pose of the source robot and reproduces it via a Cartesian impedance controller. The demonstration data was smoothed to mitigate the noise arising from the teleoperation setup. Each observation $\vo^s$ is composed of the end-effector position and of the image embedding obtained from the ResNet vision backbone that processes the images from an over-the-shoulder camera. The policies are trained to generate $8$-dimensional actions composed of the position, orientation, and gripper state lying on the product of manifolds $\euclideanspace^3 \times \mathcal{S}^3 \times \euclideanspace$.\looseness-1

\subsubsection{Pick \& Place}
\begin{figure}[tbp]
    \centering
    \includegraphics[width=0.32\linewidth]{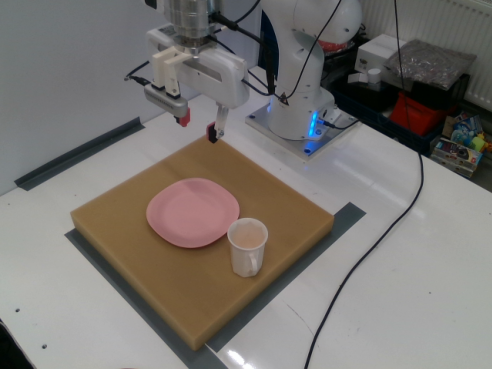}
    \hfill
    \includegraphics[width=0.32\linewidth]{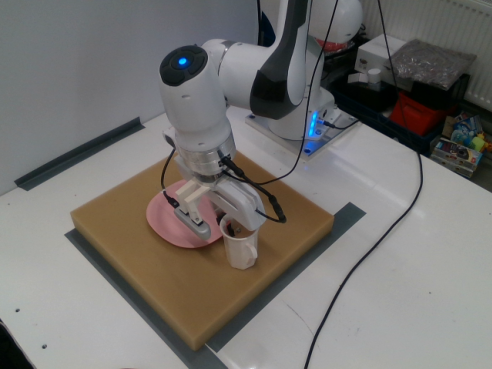}
    \hfill
    \includegraphics[width=0.32\linewidth]{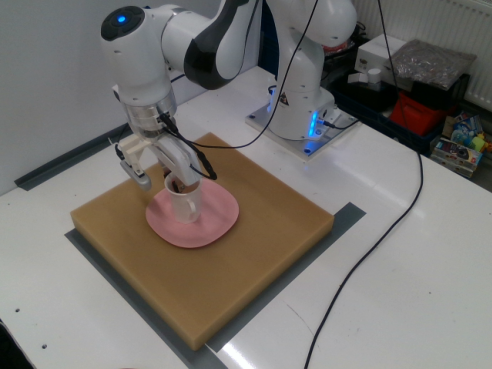}
    \caption{\textsc{Pick \& Place}: First, the robot end-effector approaches and grasps the white mug. Then, it lifts the mug and places it upright on the pink plate.}
    \label{fig:pickplace}
\end{figure}

The goal of this task is to test the ability of RFMP and SRFMP to learn Euclidean policies in real-world settings. The task consists of approaching and picking up a white mug, and place it on a pink plate, as shown in Figure~\ref{fig:pickplace}. Note that the robot end-effector points downwards during the entire task, so that its orientation remains almost constant. 

We collect $100$ demonstrations where the white mug is randomly placed on the yellow mat, while the pink plate position and end-effector initial position are slightly varied. We split our demonstration data to use $90$ demonstrations for training and $10$ for validation. All models are trained for $300$ epochs with the same training hyperparameters as reported in Table~\ref{table:hyperParamRFMP}. As in previous experiments, we use the best-performing checkpoints of each model for evaluation.

During testing, we systematically place the white mug at $10$ different locations on a semi-grid covering the surface of the yellow sponge. We evaluate the performance of RFMP, SRFMP, and DP as a function of different NFE values, under two metrics: Success rate and prediction smoothness. 
Figure~\ref{fig:real robot result} shows the increased robustness of RFMP and SRFMP to NFE compared to DP. Notably, DP requires more NFEs to achieve a success rate competitive to RFMP and SRFMP, which display high performance with only $2$ NFE. Moreover, DP generated highly jerky predictions when using $2$ NFE. In contrast, RFMP and SRFMP consistently retrieve smooth trajectories, regardless of the NFE.

Notice that the experiments involved natural variations in background and lighting conditions, further exposing the policies to realistic sensor noise. These slight variations in background and lighting conditions (e.g., cloudy and sunny days), had minimal impact on the policies performance. However, consistent failures were observed when the mug was initially positioned on the right side of the yellow mat, where the robot often obstructs the external camera view when approaching the mug. A multi-camera setting may improve performance on such occlusion cases. 


\subsubsection{Mug Flipping}
In this task, a white mug is initially positioned horizontally on a yellow sponge, as shown in Figure~\ref{fig:rotatemug}. The task consists of two stages: First, the robot locates the white mug and grasps it by rotating its end-effector to align with the mug orientation. The robot then places the mug upright on the blue plate. Note that this task demands the robot to execute elaborated rotation trajectories for both grasping and placing. For this task, we collect $50$ demonstrations with the white mug randomly positioned and rotated on the left side of a yellow sponge. Note that we use only the left side as the task requires the robot to operate near its workspace limits, which are prohibitive when the mug is placed on the right side. Furthermore, the end-effector initial pose was also slightly varied across the demonstrations. The policy hyperparameters for this task are provided in Table~\ref{table:hyperParamRFMP}.

\begin{figure}[tbp]
\centering
\vspace{-0.4cm}
\includegraphics[width=\linewidth]{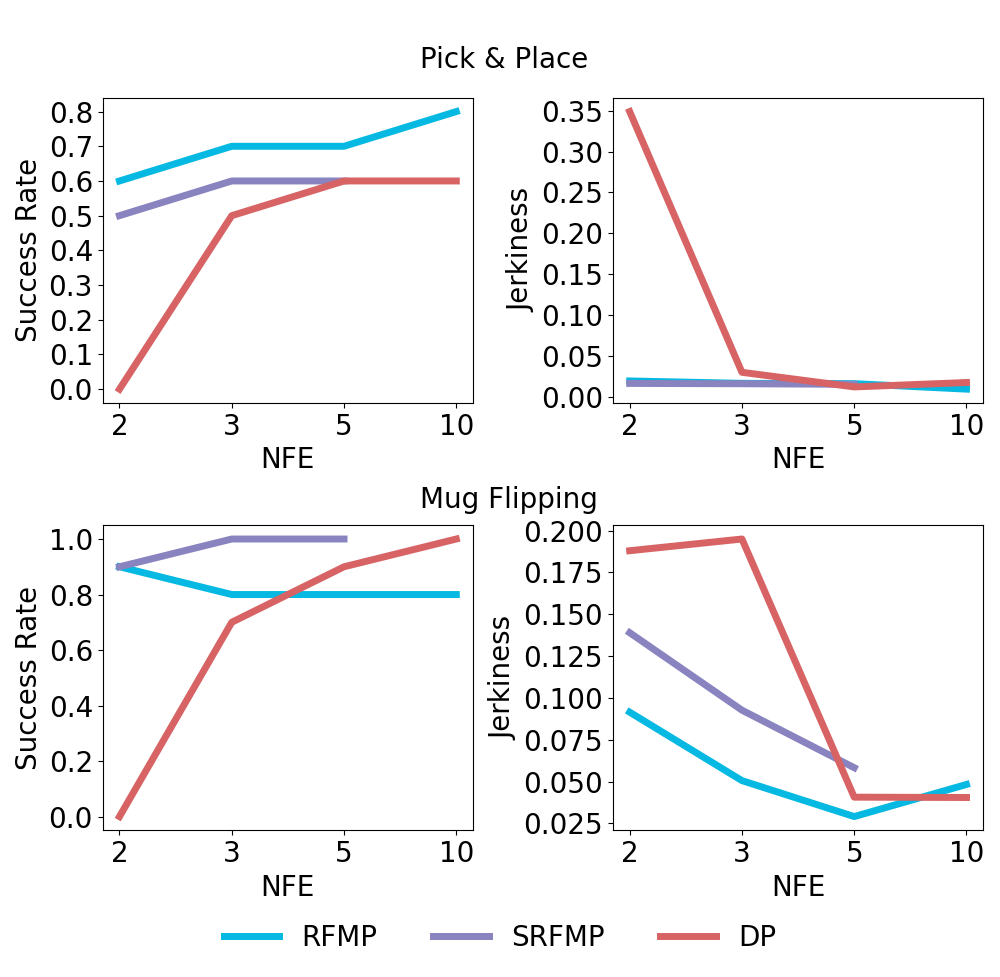}
\caption{Success rate and predicted actions jerkiness as a function of NFE on the \textsc{Pick \& Place} and \textsc{Mug flipping} tasks.}
\label{fig:real robot result}
\end{figure}

Figure~\ref{fig:real robot result} shows the results of evaluating the different considered policies using the same two metrics as the \textsc{Pick \& Place} task, namely success rate and trajectory smoothness. Both RFMP and SRFMP are significantly more robust to different NFE in terms of success rate when compared to DP. While the smoothness of RFMP and SRFMP is slightly affected by NFE in this particular task, both methods still outperform DP in this regard. Similarly to the \textsc{Pick \& Place}, slight variations on lightning and background had a negligible effect on the performance of the tested policies.

\subsection{Experiments Summary}
We conducted comprehensive evaluations across diverse benchmarks in simulation, namely the Euclidean and \textsc{Sphere Push-T} tasks, the Robomimic benchmark under both state- and vision-based observation, the long-horizon \textsc{Franka Kitchen} benchmark, as well as two real-world manipulation tasks.
These tasks display various action spaces with different types of geometric constraints. 
While the Euclidean \textsc{\mbox{Push-T}} and Robomimic tasks are defined on standard Euclidean spaces --- a special type of Riemannian manifolds --- the \textsc{Sphere Push-T} task involves an action space constrained to the hypersphere $\mathcal{S}^2$, and both the \textsc{Franka Kitchen} and real-world experiments operate on product Euclidean and sphere manifolds, accounting for position, orientation, and gripper state. Through their Riemannian formulation, RFMP and SRFMP are designed to naturally handle Euclidean and non-Euclidean action spaces, thus satisfying constraints such as the unit-norm of quaternions. 
It is important to emphasize that the Riemannian formulation is necessary not only to handle geometric constraints but also to guarantee stability in SRFMP. Naive approximations, e.g., unit-norm normalization as a post-processing step for quaternions, would break the stability guarantees in SRFMP. 

Our findings from both simulated and real-world tasks show that RFMP and SRFMP offer significant advantages over DP. In particular, RFMP and SRFMP achieve faster inference by using fewer NFE without compromising success rates regardless of the observation type. This translates into highly-robust visuomotor policies. Importantly, these advantages do not come at the cost of elaborated training strategies like those used in consistency-based models. In fact, our models outperformed CP, while being notably easier and more efficient to train. Regarding the difference between SRFMP and RFMP, the results did not show significant performance gains in terms of success rate and prediction smoothness, at the exception of the long-horizon task of the \textsc{Franka kitchen} benchmark, where SRFMP significantly outperformed RFMP and DP. Nevertheless, SRFMP shows to be easier to train, achieving higher success rate than RFMP for fewer training epochs (e.g., in \textsc{Lift}, \textsc{Can}, \textsc{Square} tasks).

\section{Conclusion}
\label{section:conclusion}
This paper introduced Stable Riemannian Flow Matching Policy (SRFMP), a novel framework that combines the easy training of flow matching with stability-based robustness properties for visuomotor policy learning.
SRFMP builds on our extension of stable flow matching to Riemannian manifolds, providing stable convergence of the learned flow to the support of Riemannian target distributions. Our simulated and real-world experiments show that both our previous work on RFMP and its stable counterpart SRFMP outperform diffusion policies and distillation-based extensions, while offering advantages in terms of  inference speed, ease of training, and robust performance even with limited NFEs and training epochs. Future work will focus on exploring equivariant policy structures to potentially reduce the number of required demonstrations and to improve generalization. Additionally, we aim to investigate multi-modal perception backbones for tackling contact-rich tasks.

\bibliographystyle{IEEEtran}
\bibliography{literature}

\vskip -2\baselineskip plus -1fil
\begin{IEEEbiography}
[{\includegraphics[width=1in,height=1.25in, clip]{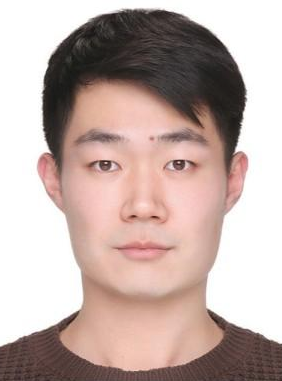}}]{Haoran Ding}
received the B.S. degree in Engineering from Tongji University, Shanghai, China, in 2021, and the M.Sc. degree in Computational Engineering from the Technical University of Darmstadt, Germany, in 2024. During his master’s studies, he worked on the robot air hockey project at the Intelligent Autonomous Systems (IAS) Group 
and completed a one-year research internship at the Bosch Center for Artificial Intelligence (BCAI), where he conducted his master thesis.  He is currently pursuing a Ph.D. degree in the Robotics Department at Mohamed bin Zayed University of Artificial Intelligence (MBZUAI), Abu Dhabi, United Arab Emirates. His research interests include visuomotor policy learning, generative models and imitation learning. 
\end{IEEEbiography}

\vfill\null

\begin{IEEEbiography}
[{\includegraphics[width=1in,height=1.25in, trim={0.4cm 0.0cm 0.4cm 0.0cm},clip]{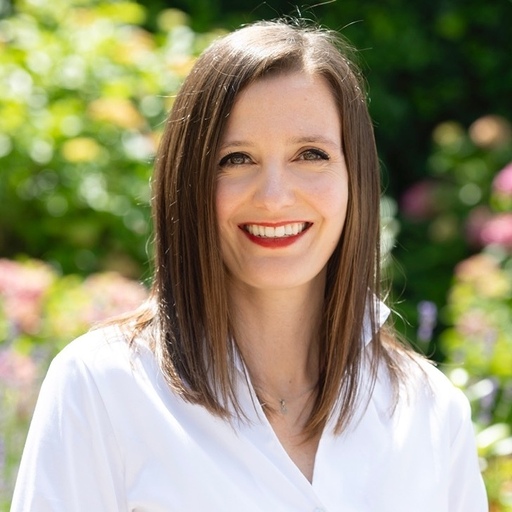}}]{Noémie Jaquier} is an assistant professor at the KTH Royal Institute of Technology, where she heads the Geometric Robot (GeoRob) Lab at the Division of Robotics, Perception and Learning. 
She received a B.Sc. degree on Microengineering, a M.Sc. degree in Robotics and Autonomous Systems from the Ecole Polytechnique Fédérale de Lausanne (EPFL) in 2014 and 2016. She received her PhD degree from EPFL in 2020 for her thesis ``Robot Skills Learning with Riemannian Manifolds: Leveraging Geometry-awareness in Robot Learning, Optimization and Control''. Prior to joining KTH, Noémie Jaquier was a postdoctoral researcher in the High Performance Humanoid Technologies Lab (H²T) at the Karlsruhe Institute of Technology (KIT) and a visiting postdoctoral scholar at the Stanford Robotics Lab. Her research investigates data-efficient and theoretically-sound learning algorithms that leverage differential geometry- and physics-based inductive bias to endow robots with close-to-human learning and adaptation capabilities. Personal webpage: \url{http://njaquier.ch}.
\end{IEEEbiography}

\vfill\null

\begin{IEEEbiography}
[{\includegraphics[width=1in,height=1.25in,trim={4cm 0.0cm 2cm 0.0cm},clip]{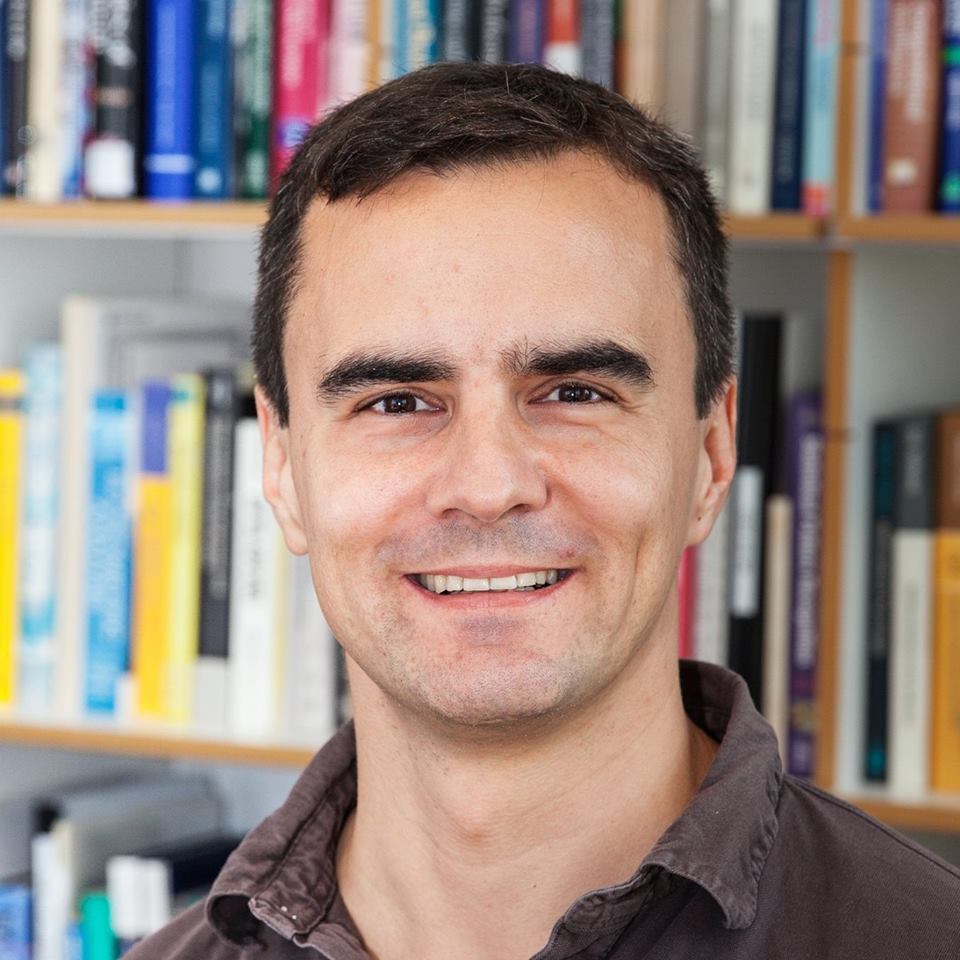}}]{Jan Peters}
is a full professor (W3) for Intelligent Autonomous Systems at the Computer Science Department of the Technische Universitaet Darmstadt since 2011, and, at the same time, he is the dept head of the research department on Systems AI for Robot Learning (SAIROL) at the German Research Center for Artificial Intelligence (Deutsches Forschungszentrum für Künstliche Intelligenz, DFKI) since 2022. He is also is a founding research faculty member of the Hessian Center for Artificial Intelligence. Jan Peters has received the Dick Volz Best 2007 US PhD Thesis Runner-Up Award, the Robotics: Science \& Systems - Early Career Spotlight, the INNS Young Investigator Award, and the IEEE Robotics \& Automation Society's Early Career Award as well as numerous best paper awards. In 2015, he received an ERC Starting Grant and in 2019, he was appointed IEEE Fellow, in 2020 ELLIS fellow and in 2021 AAIA fellow.
Jan Peters has studied Computer Science, Electrical, Mechanical and Control Engineering at TU Munich and FernUni Hagen in Germany, at the National University of Singapore (NUS) and the University of Southern California (USC). He has received four Master's degrees in these disciplines as well as a Computer Science PhD from USC. Jan Peters has performed research in Germany at DLR, TU Munich and the Max Planck Institute for Biological Cybernetics (in addition to the institutions above), in Japan at the Advanced Telecommunication Research Center (ATR), at USC and at both NUS and Siemens Advanced Engineering in Singapore. He has led research groups on Machine Learning for Robotics at the Max Planck Institutes for Biological Cybernetics (2007-2010) and Intelligent Systems (2010-2021).
\end{IEEEbiography}

\vfill\null

\begin{IEEEbiography}
[{\includegraphics[width=1in,height=1.25in,trim={2.0cm 0.0cm 2.1cm 0.0cm},clip]{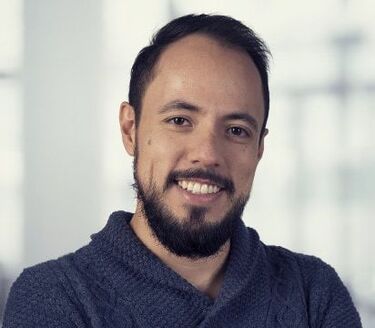}}]{Leonel Rozo} is a lead research scientist at the Bosch Center for Artificial Intelligence (BCAI), Germany. He received a Bachelor degree in Mechatronics Engineering from the ``Nueva Granada'' Military University in Bogotá, Colombia in 2005, and a Master degree in Automatic Control and Robotics from the Polytechnical University of Catalonia, Barcelona, Spain, in 2007. He carried out his PhD research at the Institut de Robòtica i Informàtica Industrial under the supervision of Prof. Carme Torras and Dr. Pablo Jiménez, and received a Ph.D in Robotics from the Polytechnical University of Catalonia in 2013. 
Prior to joining BCAI in 2018, Leonel Rozo led the Learning and Interaction Group at the department of Advanced Robotics in the Italian Institute of Technology (IIT) from 2016 to 2018. He previously joined IIT in 2012, first as a research fellow and then as postdoctoral researcher in 2013. In 2017 he was awarded an individual Marie Skłodowska-Curie fellowship for his project proposal DRAPer.
His research has been mainly focused on exploiting machine learning techniques, optimal control, and Riemannian manifold theory to learn robot motion skills via human demonstrations and refinement methods such as reinforcement learning with applications to (dual-arm) manipulation tasks and human-robot collaboration. Personal webpage: \url{https://leonelrozo.weebly.com}.
\end{IEEEbiography}

\end{document}